\documentclass[12pt]{spieman}  % 12pt font required by SPIE;
\usepackage{amsmath,amsfonts,amssymb}
\usepackage{graphicx}
\usepackage{setspace}
\usepackage{tocloft}
\usepackage{lineno}

\usepackage{amssymb}
\usepackage{multirow}
\usepackage{arydshln}
\usepackage{subcaption}

\usepackage{bm}

\def\app#1#2{%
  \mathrel{%
    \setbox0=\hbox{$#1\sim$}%
    \setbox2=\hbox{%
      \rlap{\hbox{$#1\propto$}}%
      \lower1.1\ht0\box0%
    }%
    \raise0.25\ht2\box2%
  }%
}

\newcommand{\R}{\mathcal{R}}

\title{Towards 3D Scene Understanding of Gas Plumes in LWIR Hyperspectral Images Using Neural Radiance Fields}

\author[a,*]{Scout Jarman}
\author[b]{Zigfried Hampel-Arias}
\author[b]{Adra Carr}
\author[a]{Kevin R. Moon}
\affil[a]{Utah State University, Department of Mathematics \& Statistics, Logan, UT USA}
\affil[b]{Los Alamos National Laboratory, Intelligence and Space Research Division, Los Alamos, NM USA}

\cftpagenumbersoff{figure}
\cftpagenumbersoff{table} 
\begin{document} 
\maketitle

\begin{abstract}
Hyperspectral images (HSI) have many applications, ranging from environmental monitoring to national security, and can be used for material detection and identification.
Longwave infrared (LWIR) HSI can be used for gas plume detection and analysis.
Oftentimes, only a few images of a scene of interest are available and are analyzed individually.
The ability to combine information from multiple images into a single, cohesive representation could enhance analysis by providing more context on the scene's geometry and spectral properties.
Neural radiance fields (NeRFs) create a latent neural representation of volumetric scene properties that enable novel-view rendering and geometry reconstruction, offering a promising avenue for hyperspectral 3D scene reconstruction.
We explore the possibility of using NeRFs to create 3D scene reconstructions from LWIR HSI and demonstrate that the model can be used for the basic downstream analysis task of gas plume detection.
The physics-based DIRSIG software suite was used to generate a synthetic multi-view LWIR HSI dataset of a simple facility with a strong sulfur hexafluoride gas plume.
Our method, built on the standard Mip-NeRF architecture, combines state-of-the-art methods for hyperspectral NeRFs and sparse-view NeRFs, along with a novel adaptive weighted MSE loss.
Our final NeRF method requires around 50\% fewer training images than the standard Mip-NeRF and achieves an average PSNR of 39.8 dB with as few as 30 training images.
Gas plume detection applied to NeRF-rendered test images using the adaptive coherence estimator achieves an average AUC of 0.821 when compared with detection masks generated from ground-truth test images.
\end{abstract}

% Include a list of up to six keywords after the abstract
\keywords{LWIR Hyperspectral Images, Neural Radiance Fields, 3D Scene Reconstruction, Sparse Views, Gas Plume Detection}

% Include email contact information for the corresponding author
{\noindent \footnotesize\textbf{*}Scout Jarman,  \linkable{scout.jarman@usu.edu} }

%%%%% SPIE guidelines say I can use single or double, and I feel single looks better
% If going back, don't forget to uncomment the \end statement at the end of the document
% \begin{spacing}{2}   % use double spacing for rest of manuscript

%%%%%%%%%%%%%%%%%%%%%%%%%%%%%%%%%%%%%%%%%%%%%%%%%%
\section{Introduction}
%%%%%%%%%%%%%%%%%%%%%%%%%%%%%%%%%%%%%%%%%%%%%%%%%%

% Hyperspectral images are becoming increasingly accessible and useful, and can be used for gas plume analysis.

Hyperspectral images (HSI) are information-dense, allowing them to be used for many tasks ranging from national security and disaster response, to medical diagnosis and environmental monitoring \cite{Bhargava2024HyperspectralIA, Yuen2010AnIT, Krekeler2023ANH, Rodrigues2022TrendsIH, Stuart2019HyperspectralII}.
Images can be captured across different wavelength ranges, and each range has different use cases, with longwave infrared (LWIR) HSI used for gas plume analysis \cite{Manolakis2019LongwaveIH}.
This is because gases can exhibit unique spectral fingerprints in LWIR, allowing them to be detected and quantified in LWIR HSI \cite{Manolakis2014LongWaveIH}.
Gas plume analysis can be divided into three steps: 1) detection, in which pixels that likely contain a gas of interest are found \cite{Messinger2004GaseousPD}; 2) identification, in which the gas is verified and false positive detects are removed \cite{Truslow2016PerformanceMF}; and 3) quantification, in which the temperature and density of the plume are calculated \cite{Turcotte2010GasPQ}.

%%%%% Gas plume detection works by ..., but analysis, particularly of plume geometry, can be limited by having a single view
% Various algorithms exist for gas plume detection, though we focus on the Adaptive Coherence Estimator (ACE) detector \cite{Manolakis2013TheRS}.
% ACE provides an estimate of the likelihood that each pixel in an image contains the signal of a given gas of interest, based on the distribution of background pixels.
% Since this is typically evaluated for a single view of the gas plume, only a rudimentary estimation of the plume's geometry can be made.
% Furthermore, the specific background underneath the plume can negatively impact detection performance and subsequently the estimation of the plume's geometry \cite{Matteoli2014AnOO, Jarman2024LocalBE}.
% Capturing multiple views of the gas plume and estimating the 3D scene geometry could greatly improve detection and subsequent gas plume analysis.

%%%%% Airborne captures typically contain only a few images. Images are most often analyzed individually. Some work has shown improved performance by using information from other photos to improve performance.
One way to collect LWIR HSI is from an airborne platform, where only a few images of a specific target can be taken.
These images are then analyzed using the three steps above to determine the presence and properties of any gas plumes in the scene.
Even if multiple images of a scene are available, it is standard practice to analyze each image individually \cite{Manolakis2014LongWaveIH, Pogorzala2005GasPS, Tremblay2010StandoffGI}.
Incorporating information from other images has been shown to improve plume analysis \cite{Gerhart2013DetectionAT}, including for detection, as the background of a plume can be better estimated \cite{Marrinan2016FlagbasedDO, Ning2025TowardsGP}.
Furthermore, single-image analysis only offers a limited understanding of the scene and plume geometry.
Using multiple views of the gas plume to estimate the full scene 3D geometry and spectral properties could greatly improve gas plume analysis.
For example, by using the shared information from multiple images to improve background estimation, or by allowing for better estimates of plume size, shape, and pathlength.

% Neural Radiance Fields offered state-of-the-art performance in neural volumetric representations of 3D scenes, and could be used to greatly improve gas plume analysis.
Neural radiance fields (NeRFs) offer state-of-the-art performance in 3D scene reconstruction and novel view rendering compared to standard photogrammetry methods \cite{Mildenhall2020NeRF}.
NeRFs represent the volumetric scene properties using neural networks.
For training, they require multiple images of the scene from different positions along with their camera locations and viewing angles.
A trained NeRF model can then predict the color and volumetric density at each point in space, allowing the rendering of unseen views and an estimate of the scene's geometry.
Some prior work has applied NeRFs to HSI with promising results; however, these approaches used data in the visible wavelength range, not LWIR \cite{Li2023SpectralNeRFPB}.
Little work has been done applying NeRFs to LWIR HSI because of the general unavailability of LWIR HSI datasets, let alone those containing a gas plume with ground truth information.

% The work presented here is the first to look at NeRFs for LWIR gas plume detection, and is an extension of the work presented in \cite{HampelArias20253DSU}, which is the first paper to look at NeRFs for LWIR HSI.
This paper is an extension of the work presented in Ref.~\citenum{HampelArias20253DSU} and Ref.~\citenum{Jarman2025HyperspectralNR}, which explored applying NeRFs to 3-channel PCA representations of LWIR HSI and adapting NeRF for full channel LWIR HSI.
Both this paper and the previous works use a simple, simulated LWIR HSI generated with the DIRSIG software suite.
This paper explores two main questions: can NeRFs be used to create a single, cohesive representation of an LWIR HSI scene, and can such a model be used to improve gas plume analysis?
Since detection is the fundamental first step in gas plume analysis, we explore basic gas plume detection in NeRF-rendered images.
This paper answers these two questions and has the following contributions:
\begin{enumerate}
    \item The combination and comparison of state-of-the-art NeRF methods, namely multi-channel density NeRF from the HSI NeRF literature \cite{Ma2025MultichannelVD} and RegNeRF's geometry regularization from the sparse-view NeRF literature \cite{Niemeyer2021RegNeRFRN}, and proposal of a three-piece loss function that includes an adaptive weighted MSE term that increases gas plume detection performance in rendered views.
    Our model outperforms each method individually and requires fewer training images.
    \item A demonstration of the possible utility of NeRF for gas plume analysis by comparing gas plume detection maps from held-out testing images, and NeRF renderings of those testing images.
    This sets up future work to explore how our model could be used to improve upon, rather than simply match, existing gas plume detection methods.
\end{enumerate}

% Outline, background, methods, results, conclusions.
This paper is structured as follows.
Section \ref{sec:Background} provides background on photogrammetry, NeRFs, and their application to HSI.
Additionally, a description of gas plume detection is given.
Section \ref{sec:Methods} describes our adaptations to the NeRF model.
Section \ref{sec:Results} details the data we used for training and the performance of our NeRF model in both novel HSI rendering as well as in gas plume detection.
Section \ref{sec:Conclusions} summarizes the results and discusses future work.
The Appendix presents an ablation study comparing our model to the NeRF models on which it is based, as well as basic timing and GPU usage statistics.

%%%%%%%%%%%%%%%%%%%%%%%%%%%%%%%%%%%%%%%%%%%%%%%%%%
\section{Background}
\label{sec:Background}
%%%%%%%%%%%%%%%%%%%%%%%%%%%%%%%%%%%%%%%%%%%%%%%%%%

% Brief history of photogrammetry, and the applications of it to HSI.

Photogrammetry is the process of using multiple images to represent a scene in three-dimensional space, including geometric and occasionally color information.
Two primary photogrammetry methods are Multi-view Stereo and Structure from Motion (SFM).
Multi-View Stereo requires camera poses and view angles, also known as camera extrinsics, to create a dense 3D point cloud from neighboring camera views \cite{Seitz2006ACA, Schnberger2016PixelwiseVS}.
SFM uses feature detection and matching to create a sparse 3D point cloud of the structures present in the scene \cite{Schnberger2016StructurefromMotionR, Westoby2012StructurefromMotionPA}.
One advantage of SFM is that it does not require camera extrinsics.

With the rise in computational power and the development of general-purpose GPUs, neural networks have made breakthroughs across many areas of research, particularly in computer vision \cite{LeCun2015DeepL, Schmidhuber2014DeepLI}.
The advent of neural networks allowed for a new approach to photogrammetry.
Instead of creating an explicit representation of the scene with a point cloud, neural rendering methods store the geometric and color information in the neural network itself \cite{Sitzmann2019SceneRN, Lombardi2019NeuralV, Sitzmann2020ImplicitNR, Kato2017Neural3M, Tulsiani2017MultiviewSF}.
NeRFs provide state-of-the-art performance in neural rendering, offering new opportunities for photogrammetry applications \cite{Mildenhall2020NeRF}.

Photogrammetry has been successfully applied to many fields, including archaeology, cartography, and medicine \cite{MarnBuzn2021PhotogrammetryAA, Reinoso2018CartographyFC, Talevi2023EvaluationOP}.
However, scenes with simple textures or complex lighting conditions can make traditional photogrammetry difficult due to limitations in feature matching and triangulation \cite{Kadhim2023ACR, Wu2012IntegratedPA}.
Photogrammetry has been applied to HSI, though the increase in data dimensionality, along with the reflectance and illumination properties of HSI, has made its application difficult and computationally expensive \cite{Miller2015HyperspectralSA, feng2023review}.
Hyperspectral images, paired with SFM applied to panchromatic images, enabled the extraction of spatial and spectral information in Ref.~\citenum{Liu20183DRF}.
SFM was directly applied to aerial hyperspectral images in Ref.~\citenum{Miller2014Passive3S}, producing a 3D surface with spectral information.
In Ref.~\citenum{Zia20153DRF}, SFM is applied per channel, and an alignment is learned to create the final 3D surface.

There has been some work in specifically reconstructing the 3D geometry of gas plumes.
In Ref.~\citenum{Hu2022ThreedimensionalRO}, the 3D concentration pathlength of a gas plume is reconstructed by capturing the plume from two HSI sensors.
The concentration pathlength is then estimated for each image, and reconstructed slices of the plume using the simultaneous algebraic reconstruction technique are stacked to form a 3D voxel representation of the plume's concentration in space.
In Ref.~\citenum{Zhang2025A3R}, two multi-spectral sensors are used to image a gas plume, and the non-axisymmetric inverse Abel transform method is used to fit the shape of the gas plume along both axes of the plume.
Ref.~\citenum{Donato2016AdvancesI3} presents a similar method for 3D plume concentration reconstruction, but using three SIGIS spectrometers over a much larger area than the previous methods.
These methods of 3D gas plume reconstruction are more akin to tomography than to SFM, and focus solely on reconstructing the plume concentration geometry rather than the spectral signatures of the scene.
 
NeRFs are being applied to HSI with promising results compared to traditional photogrammetry methods.
A comparison of NeRF architectures applied to three-channel spectral images was done in Ref.~\citenum{Feng2024Hyperspectral3R}.
A wavelength positional encoding, allowing for spectral super-resolution, was applied to the nerfacto model from nerfstudio \cite{Tancik2023NerfstudioAM} in Ref.~\citenum{Chen2024HyperspectralNR}.
In Ref.~\citenum{Ma2024HyperspectralNR}, an inductive bias is used -- having the network predict the difference from a given reference signature -- allowing for the generation of images with different illumination properties.
In Ref.~\citenum{Ma2025MultichannelVD}, individual densities are learned for each channel, instead of a single density for all wavelengths, leading to increased image reconstruction performance.
In Ref.~\citenum{HampelArias20253DSU} and Ref.~\citenum{Jarman2025HyperspectralNR}, NeRFs were applied to LWIR HSI for the first time by first using PCA to reduce the spectral images to three channels, then applying existing NeRF architectures.
Our work is a direct extension of these and aims to further improve performance with sparse training views of full spectral signatures and to explore the application of NeRFs to downstream gas plume analysis tasks.

Further research has also been conducted to improve NeRFs for few-shot learning and sparse-view datasets.
Adapting these works to HSI NeRFs is important, as HSI data is often more difficult to obtain, leaving only a few HSI captures available in practice.
Many works rely on using external information for additional supervision for the NeRF.
In Ref.~\citenum{Deng2021DepthsupervisedNF}, SFM is used to estimate an initial set of depth points, which can be used to supervise the NeRF depth predictions.
In Ref.~\citenum{Jain2021PuttingNO}, random patches are regularized against CLIP embeddings to ensure that novel views are semantically consistent.
However, SFM is difficult to apply to HSI, particularly for relatively simplistic scenes, and models like CLIP are trained on millions of RGB images, which is not possible for HSI.
A promising sparse-view NeRF solution is RegNeRF, proposed in Ref.~\citenum{Niemeyer2021RegNeRFRN}.
Here, random patches are regularized to be piecewise smooth, requiring no supervision or pretraining on a large image library.
Our work is the first to adapt RegNeRF for use with HSI and is adapted to multi-channel density NeRF.

\subsection{NeRF Technical Details}

NeRFs enable novel-view rendering by representing a 3D scene as a continuous volumetric radiance field stored latently in a neural network \cite{Mildenhall2020NeRF}.
The network learns a function that maps a five-dimensional input to a color $\mathbf{c}(\mathbf{x}, \mathbf{d})\in\mathbb{R}^3$ and volume density $\sigma(\mathbf{x})>0$, where the inputs are a 3D position $\mathbf{x}\in\mathbb{R}^3$ and 2D viewing direction $\mathbf{d}\in\R^2$.
The network typically consists of a base multi-layer perceptron (MLP) with about eight layers to predict density, and a branched four-layer MLP to predict color.
Individual pixels can then be rendered with volumetric rendering using samples along the ray cast from the pixel.
The ray for a pixel is defined as $\mathbf{r}(t)=\mathbf{o}+t\mathbf{d}$, where $\mathbf{o}$ is the ray origin and $t$ is the distance along the ray.
The final rendered color $\hat{\mathbf{C}}(\mathbf{r})$ is found by sampling the ray $N$ times as
\begin{equation}
    \label{eq:volume render}
    \begin{tabular}{c}
        $\hat{\mathbf{C}}(\mathbf{r}) = \sum_{i=1}^N T_i\alpha_i\mathbf{c}_i,$\\\\
        $\alpha_i=1-\exp(-\sigma_i \delta_i), \quad T_i=\exp\left(-\sum_{j=1}^{i-1}\sigma_j\delta_j\right),$
    \end{tabular}
\end{equation}
where $\delta_i=t_{i+1}-t_i$ is the interval between samples.
This rendering process is fully differentiable, allowing the network to learn the scene's volumetric field.

The NeRF samples $\mathbf{r}$ in two passes, which is called hierarchical volume sampling \cite{Mildenhall2020NeRF}.
First, the ray is uniformly sampled $N_c$ times between predefined near- and far-planes.
A ``coarse" color prediction is then calculated as $\hat{\mathbf{C}}_c(\mathbf{r})=\sum_{i=1}^{N_c}T_i\alpha_i \mathbf{c}_i$.
These samples are then used to estimate the distribution of densities along the ray, such that areas with higher densities can be more sampled.
This PDF is sampled $N_f$ times using inverse transform sampling based on the weights $w_i=T_i\alpha_i$, and is normalized to sum to one.
The final ``fine" color prediction $\hat{\mathbf{C}}_f(\mathbf{r})$ is then calculated by Equation \ref{eq:volume render} using all $N=N_c+N_f$ samples.
Note that we refer to the final fine color rendering as $\hat{\mathbf{C}}(\mathbf{r})$.

Typically, a standard L2 loss function is used to measure the error between the rendered and true pixels.
Note that both the coarse and fine color renderings are used in the loss function:
\begin{equation}
\label{eq:L2}
    \mathcal{L}_{L2}=\lambda_c||\hat{\mathbf{C}}_c(\mathbf{r})-\mathbf{C}(\mathbf{r})||^2_2 + \lambda_f||\hat{\mathbf{C}}_f(\mathbf{r})-\mathbf{C}(\mathbf{r})||^2_2,
\end{equation}
where $\hat{\mathbf{C}}_c$ is coarse color rendering, $\hat{\mathbf{C}}_f$ is the fine color rendering, $\mathbf{C}$ is the ground truth color, and $\lambda_c$ and $\lambda_f$ control how much each term contributes to the full loss.
We use $\lambda_c=0.1$ and $\lambda_f=1$.

NeRFs use simple MLPs to learn the mapping of position and direction to color and density.
However, MLPs are biased towards low-frequency features, meaning they struggle to learn sharp or high-frequency details in scene geometry and lighting conditions \cite{Rahaman2018OnTS, Xu2022OverviewFP}.
One solution to help the network learn fine details is to use positional encoding, which maps input features to a higher-dimensional space using sinusoidal functions \cite{Tancik2020FourierFL}.
For example, the $y$ coordinate would be mapped to
\begin{equation}
    \gamma(y)=[\sin(2^0\pi y), \cos(2^0\pi y), ..., \sin(2^{L-1}\pi y), \cos(2^{L-1}\pi y)],
\end{equation}
where $L$ controls the size of the feature space, and provides for varied input feature frequencies.
Positional encoding is applied to both input coordinates $\mathbf{x}$ and viewing direction $\mathbf{d}$.
The original NeRF implementation uses $L=10$ for the spatial input, and $L=4$ for the viewing direction.

\subsection{Gas Plume Detection}
\label{sec:detection}

Gas plume analysis using hyperspectral images often involves three common steps.
\begin{enumerate}
    \item Detection: determine which pixels likely contain a gas of interest \cite{Manolakis2014DetectionAI, Manolakis2002DetectionAF}
    \item Identification: verify plume substance and reduce false positive detections \cite{Truslow2016PerformanceMF, Klein2023HyperspectralTI, Jarman2025ImprovedGP}
    \item Quantification: determine density, temperature, and other characteristics of the plume \cite{Idoughi2016BackgroundRE, Hayden1997RemoteTG}
\end{enumerate}
If NeRFs are able to sufficiently and accurately reconstruct plume geometry and spectral features of a plume, all of these plume analysis steps could be improved.
For example, 3D scene understanding could lead to improved background estimation, greatly improving the ability to detect and quantify a plume.
Additionally, the plume's full 3D geometry could be extracted, along with 3D estimates of plume temperature and concentration.
We focus on applying gas plume detection to novel rendered NeRF views, since all subsequent plume analyses rely on this first step.

While various gas plume detection algorithms exist, we use the adaptive coherence estimator (ACE) detector because it offers strong out-of-the-box detection performance \cite{Manolakis2013TheRS}.
To derive the ACE detector, the additive spectral model is used, which assumes that any spectral signature can be modeled as
\begin{equation}
    \mathbf{y}=\mathbf{b}+a\mathbf{t}, \quad \mathbf{b}\sim N(\mathbf{\mu}, \Sigma),
\end{equation}
where $\mathbf{y}$ is the observed spectral signature at a given pixel, $\mathbf{b}$ is the background signature assumed to be multivariate-normally distributed, $\mathbf{t}$ is the target signature of interest, and $a$ is the strength of the target signature.
We wish to test whether the gas is present in the pixel ($a\neq0$) or absent ($a=0$).
The solution to the hypothesis test $H_0:a=0, H_A:a\neq0$ using the generalized likelihood ratio test results in the ACE score being
\begin{equation}
\label{eq:ace}
    ACE(\mathbf{y}) = \frac{\left(\mathbf{t}^T\Sigma^{-1}(\mathbf{y}-\mu)^T\right)^2}{(\mathbf{t}^T\Sigma^{-1}\mathbf{t})\cdot \left((\mathbf{y}-\mu)^T\Sigma^{-1}(\mathbf{y}-\mu)\right)}.
\end{equation}

\begin{figure*}[]
    \centering
    \includegraphics[width=\linewidth]{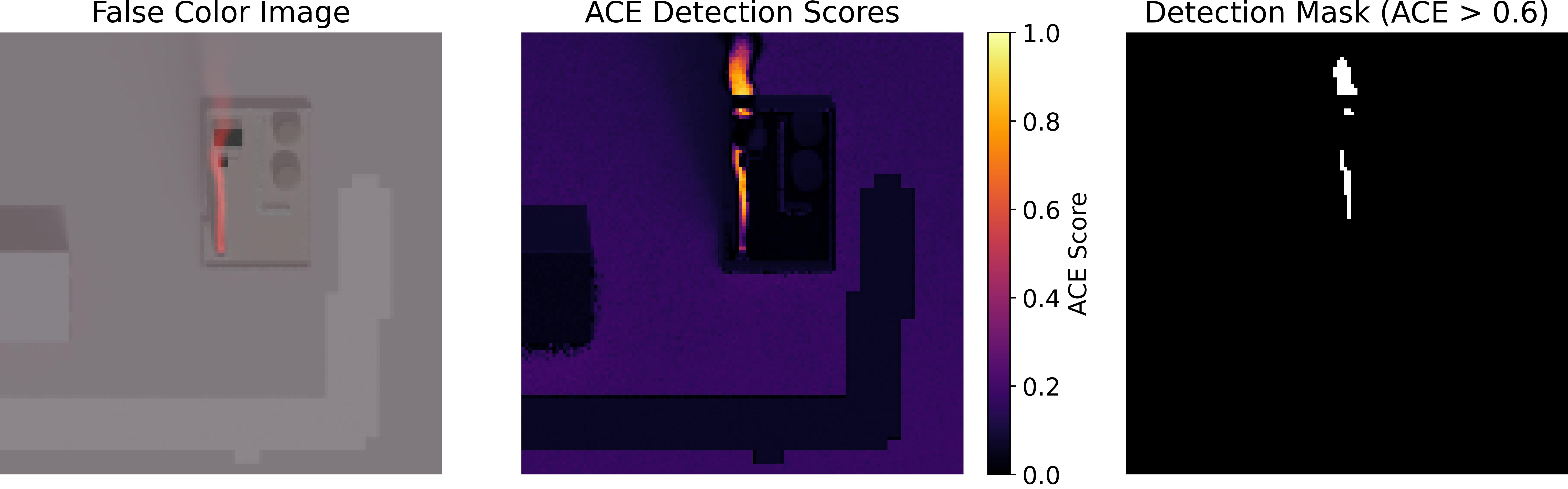}
    \caption{Example of the gas plume detection process. (Left): False-coloringing HSI using wavelengths 10.4, 8.1, and 8.5 $\mu$m for the red, green, and blue channels. (Middle): ACE detection score map with background $\mu$ and $\Sigma$ estimated using all pixels, and using the SF$_6$ absorption spectrum for $\mathbf{t}$. (Right): Plume mask created by thresholding the ACE scores at $0.6$.}
    \label{fig:ACE example}
\end{figure*}

In practice, the background distribution is estimated using all available pixels, and ACE is applied to every pixel in the scene.
Figure \ref{fig:ACE example} shows an example of the gas plume detection process.
Once the ACE scores have been computed for every pixel, a threshold is applied to create a mask representing the plume.
However, despite ACE's robustness, it can produce false positives and negatives.
For example, false negatives can be seen in the gaps in the plume detection mask in Figure \ref{fig:ACE example}.
This is an area where NeRFs may be of use.
By imaging the scene from multiple angles, a more accurate understanding of the plume's geometry and substance may be obtained.

%%%%%%%%%%%%%%%%%%%%%%%%%%%%%%%%%%%%%%%%%%%%%%%%%%
\section{Methods}
\label{sec:Methods}
%%%%%%%%%%%%%%%%%%%%%%%%%%%%%%%%%%%%%%%%%%%%%%%%%%

NeRF variants exist that offer improvements in performance over the original NeRF design.
One such variant is Mip-NeRF, which improves anti-aliasing performance compared to the original NeRF \cite{Barron2021MipNeRFAM}.
It accomplishes this by representing pixels as conical frustums rather than infinitesimal rays, enabling the model to encode the volume they occupy.
This is implemented by using a different encoding scheme, called integrated positional encoding.
This effectively allows the model to represent the scene at different resolutions, improving anti-aliasing performance when rendering zoomed-in or super-resolution images.
We chose this variant for two reasons.
First, Mip-NeRF is the model used in RegNeRF, which is a few-shot learning NeRF architecture that we implement here for HSI.
Second, our HSIs are low-resolution, with only $128\times128$ pixels, so we anticipate that improved anti-aliasing will be beneficial for rendering new perspectives.

\subsection{Loss Function}

The first alteration we make to the NeRF training process is optimizing the loss function.
The default loss used between the ground truth pixels and the volumetrically rendered pixels is the L2 loss as in Equation \ref{eq:L2}.
However, spectral shape is a very important factor in HSI and gas plume analysis, and different distance metrics can offer advantages compared to the L2 distance measure alone \cite{Deborah2015ACE}.
One such distance metric is the Spectral Angle Mapper (SAM), defined as
\begin{equation}
    \mathcal{L}_{SAM} = \cos^{-1}\left(\frac{ \hat{\mathbf{C}}(\mathbf{r})^T \mathbf{C}(\mathbf{r})}{||\hat{\mathbf{C}}(\mathbf{r})||_2\cdot ||\mathbf{C}(\mathbf{r})||_2}\right),
\end{equation}
where $||\cdot||_2$ is the L2 norm \cite{Yuhas1992DiscriminationAS}.
Using SAM as an additional loss function for implicit neural representation models is shown in Ref.~\citenum{Liu2023HyperspectralRS}.
SAM encourages the NeRF to learn to render pixels with spectral signatures whose shapes and correlation structures closely match those of the ground truth signatures.
The L2 loss function is still included to ensure the radiance values match the ground truth, since SAM is insensitive to scaling and magnitude.

During our experiments, we noticed that the NeRF struggled with certain wavelength ranges more than others.
Specifically, the wavelengths that correspond to the gas plume tended to have larger errors and uncertainties.
We investigated adding a weighted L2 loss function, defined as
\begin{equation}
    \label{eq:AWL2}
    \mathcal{L}_{WL2}=\sum_{j=1}^c w_j (\hat{\mathbf{C}}(\mathbf{r})_j-\mathbf{C}(\mathbf{r})_j)^2,
\end{equation}
where $w_j$ is the weighting for the $j^\text{th}$ channel, and $c$ is the number of spectral channels.
We also included the constraint that $w_j\geq0$ and $\sum_{j=1}^c w_j=1$.
A common weighting scheme is to base the weights on the channel variances \cite{Liu2023HyperspectralRS, Li2023SpectralNeRFPB}.
However, since the errors occurred at wavelengths associated with the gas, channel-variance weighting likely would not help.
An option is to use the absorption spectra $\mathbf{t}$ of the gas of interest as a weighting.
Though without knowing the gas \textit{a priori}, this could introduce bias to the NeRF.

\begin{figure*}[]
    \centering
    \includegraphics[width=\linewidth]{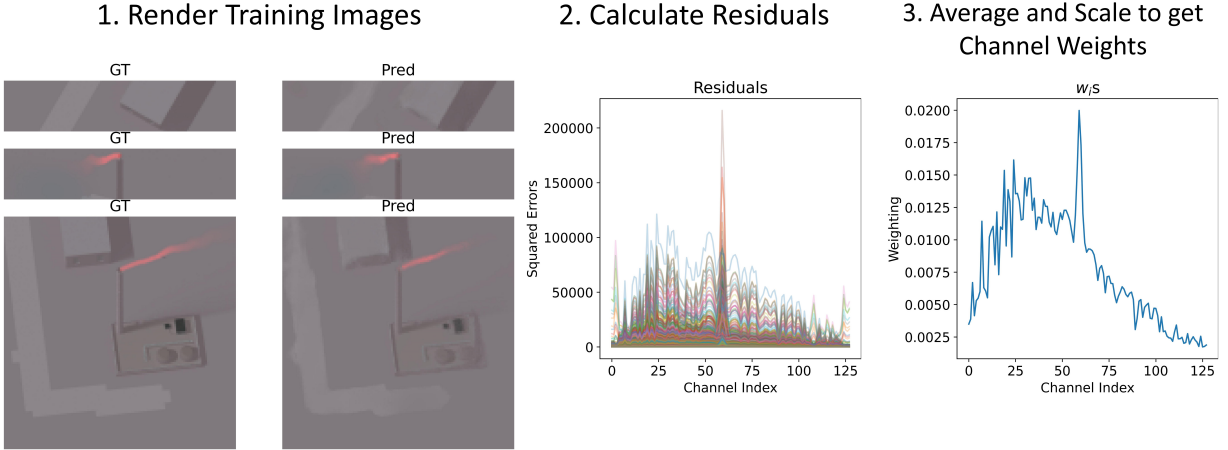}
    \caption{Example of the adaptive weighted L2 loss weight calculation. 1. provides example false-colored training images with their associated NeRF renderings. 2. shows the squared channel residuals for each pixel. 3. shows the weighting for each channel after averaging the residuals and scaling to sum to one.}
    \label{fig:awl2}
\end{figure*}

Assuming no \textit{a priori} information, we developed an adaptive weighting scheme based on the model's residuals.
Figure \ref{fig:awl2} shows an example illustrating how the weights are defined.
The model first generates predictions for each training pixel, and then calculates squared residuals.
The average squared residual for each channel is then used as the weighting:
\begin{equation}
    w_j = \frac{1}{\mathcal{N}} \sum^\mathcal{N} \left(\hat{\mathbf{C}}(\mathbf{r})_j - \mathbf{C}(\mathbf{r})_j\right)^2,
\end{equation}
where $\mathcal{N}$ is the total number of training pixels.
The weights are normalized to sum to one before being used in Equation \ref{eq:AWL2}.
We call this adaptive weighted L2 loss (AWL2) since these weights can be updated every $N$ iterations, allowing the model to adjust channel importance as it learns.
We update the weights every 5000 iterations.

Our final spectral loss function is
\begin{equation}
    \mathcal{L}_{spec}=\mathcal{L}_{L2} + \lambda_{SAM}\mathcal{L}_{SAM} + \lambda_{AWL2}\mathcal{L}_{AWL2}.
\end{equation}
The regularization parameters for each term were determined empirically, with $\lambda_{SAM}=2$.
Since $\mathcal{L}_{AWL2}$ requires model predictions, the model must first train for enough time to begin to recreate reasonable spectral signatures.
For this reason, we schedule $\lambda_{AWL2}$ such that it starts at 0, then linearly increases to 100 between iterations 5000 and 25000, and remains at 100 for the remainder of training.
It should be noted that though $\mathcal{L}_{L2}$ uses both the coarse and fine color renderings, $\mathcal{L}_{SAM}$ and $\mathcal{L}_{AWL2}$ are only applied to the fine color rendering.

\subsection{HSI NeRF Options}

We now consider various alterations to NeRF from the HSI NeRF literature.
We first consider a modification of an implicit bias as proposed in Ref.~\citenum{Ma2024HyperspectralNR}, where the NeRF is trained to predict the residual from a given reference signature instead of the original signatures themselves.
This allowed the authors to simulate different illumination conditions by changing the reference signature.
We achieve a similar effect by standardizing our training data, where the channel means and standard deviations are estimated from the training pixels.
This effectively results in the NeRF predicting scaled residuals from the reference global background signature.
It has also been shown that standardizing data for use in neural networks generally improves performance \cite{Shanker1996EffectOD}.

The next change we must make is to the NeRF architecture to accommodate the fact that HSI images have more than three channels, unlike RGB images.
One could apply dimensionality reduction to reduce images to three channels, allowing the use of all existing color models \cite{HampelArias20253DSU}.
However, we are interested in preserving all available spectral information and thus choose not to pursue this option.
Another option is to simply increase the color output dimension from three to the number of spectral channels (128 for our images) \cite{Li2023SpectralNeRFPB, Ma2024HyperspectralNR}.
Another option is to positionally encode wavelength as an input, which can allow for spectral super-resolution \cite{Chen2024HyperspectralNR}.
However, as we are not interested in spectral super-solution, we opt for the simpler solution of having the NeRF output 128 color channels.

Lastly, we consider the number of density outputs.
By default, NeRF will output a single density $\sigma$ for a given location.
However, in Ref.~\citenum{Ma2025MultichannelVD}, the NeRF predicts a separate density for each spectral channel, leading to increased image reconstruction performance.
This multi-channel density (MD) approach has the potential to align with the intuition of gas plume phenomenology, where a gas is effectively invisible at wavelengths it does not absorb.
Allowing the NeRF to learn a density for each channel will enable it to learn the plume's density at the wavelengths it absorbs.

Using MD slightly changes how the final color is rendered.
Each spectral channel is rendered as 
\begin{equation}
    \begin{tabular}{c}
        $\hat{\mathbf{C}}(\mathbf{r})_j = \sum_{i=1}^N T_{ij}\alpha_{ij} \mathbf{c}_{ij},$\\\\
        $T_{ij} = \exp\left(-\sum_{k=1}^{i-1}\sigma_{kj}\delta_k\right),\quad \alpha_{ij}=1-\exp(-\sigma_{ij}\delta_i),$
    \end{tabular}
\end{equation}
where $\mathbf{c}_{ij}$ is the predicted radiance for the $j^\text{th}$ spectral channel of the $i^\text{th}$ ray sample.
Using MD also alters the PDF weight calculation for the additional fine ray samples.
Individual channel ray weights are calculated as $w_{ij}=T_{ij}\alpha_{ij}$, and the final PDF weights are calculated as the sum of the per channel weights $w_{i}^* = \sum_{j=1}^c w_{ij}$, after normalizing to sum to one.

\subsection{Sparse View NeRF Options}

Since HSI data are generally more difficult to obtain than RGB images, in practice, there may be few HSI images available to train a NeRF.
For that reason, we consider improvements to our NeRF model from the few-shot and sparse-view NeRF literature.
Many existing methods rely on pretraining and learning semantic representations from large, readily available unlabeled image datasets \cite{Yu2020pixelNeRFNR, Chibane2021StereoRF, Chen2021MVSNeRFFG, Deng2021DepthsupervisedNF, Jain2021PuttingNO}.
However, there are very few large HSI datasets available for this purpose.
Furthermore, each sensor captures different wavelengths of light, thus any large pre-collected datasets would have to be from the same hyperspectral sensor.

For this reason, we consider the methodology applied by RegNeRF \cite{Niemeyer2021RegNeRFRN}.
RegNeRF proposes three changes to the standard Mip-NeRF, primarily by adding random patch-based regularization terms.
In addition to the supervised training samples, RegNeRF will randomly generate patches from the scene that are not present in the training data.
These random patches are generated by randomly sampling camera extrinsics within the bounding box of the training views, with slight variations in viewing angles.
The main regularization applied using these patches is geometry regularization (GR).
The assumption of GR is that the scene geometry should be piecewise smooth, which can be quantified for a patch by measuring the difference in the volumetrically rendered depth between neighboring pixels.

Similar to rendering color, the expected depth of a pixel is calculated as
\begin{equation}
    d(\mathbf{r}) = \sum_{i=1}^N T_i\alpha_i \sigma_i.
\end{equation}
However, with MD, there is not just a single $\sigma$ but a vector of densities.
Therefore, we need to adapt the existing GR to work with a vector of densities.
To do this, we first render the expected depth for each channel, indexed by $j$, as
\begin{equation}
    d_j(\mathbf{r}) = \sum_{i=1}^N T_{ij}\alpha_{ij} \sigma_{ij}.
\end{equation}
We then calculate the average depth as 
\begin{equation}
    \bar{d}(\mathbf{r}) = \frac{1}{c}\sum_{j=1}^c d_j(\mathbf{r}).
\end{equation}
Using the average depth gives equal weighting to the expected depth of each channel, though other weighting schemes could be used.
The geometry regularization for a given patch is then
\begin{equation}
    \displaystyle \mathcal{L}_{GR}= \sum_{\mathbf{r}\in\mathcal{R}_p}\sum_{i,j=1}^{S_p-1} \left(\bar{d}(\mathbf{r}_{ij}) - \bar{d}(\mathbf{r}_{i+1j})\right)^2 + \left(\bar{d}(\mathbf{r}_{ij}) - \bar{d}(\mathbf{r}_{ij+1})\right)^2
\end{equation}
where $\mathcal{R}_p$ is a randomly generated patch of pixels, $\mathbf{r}_{ij}$ is the ray at coordinate $i, j$ relative to the center ray $\mathbf{r}$, and $S_p$ is the size of the patch.  

The full loss function used to train our NeRF model at each iteration is then 
\begin{equation}
    \mathcal{L} = \mathcal{L}_{spec} + \lambda_{GR}\mathcal{L}_{GR}.
\end{equation}
We again schedule $\lambda_{GR}$ to start at $10,000$ and linearly decay to $1$ by iteration $6000$, where it remains for the rest of the training.
This strong initial regularization forces the network to produce smooth geometry across the full scene, then lets the NeRF focus on accurately modeling spectral signatures.

Another component added by RegNeRF is sample space annealing.
Sample space annealing constrains the effective near- and far-planes from which the rays can be sampled.
This leads to more consistent model convergence in early training \cite{Niemeyer2021RegNeRFRN}.
For given default near- and far-planes $t_n$ and $t_f$, the annealed sample space planes for the $k^\text{th}$ iteration are defined as 
\begin{equation}
    \begin{tabular}{c}
        $t_n(k) = t_m + (t_n-t_m)\eta(k)$\\
        $t_f(k) = t_m + (t_f-t_m)\eta(k)$\\
        $\eta(k) = \max(\min(k/N_k, p_s), 1),$
    \end{tabular}
\end{equation}
where $t_m$ is a center point between $t_n$ and $t_f$, $N_k$ is a hyperparameter controlling how long to apply annealing, and $p_s$ is a hyperparameter indicating the initial start range.
We used the midpoint between $t_n$ and $t_f$ for $t_m$, and $p_s=0.85$ to not overly restrict the sample space.
However, for remote sensing, where scene geometry tends to be closer to the center of the scene, a midpoint farther from the camera could be chosen.

Each proposed variation to the Mip-NeRF architecture led to improvements in performance, particularly with fewer training images.
An ablation study showing how each addition improves performance is given in the Appendix.
Additionally, a description of other hyperparameters and training considerations is provided.

%%%%%%%%%%%%%%%%%%%%%%%%%%%%%%%%%%%%%%%%%%%%%%%%%%
\section{Results}
\label{sec:Results}
%%%%%%%%%%%%%%%%%%%%%%%%%%%%%%%%%%%%%%%%%%%%%%%%%%

There does not exist, to the best of our knowledge, a publicly available multi-view LWIR HSI dataset of a gas plume.
For this reason, we turn to simulation; specifically, we use the Digital Imaging and Remote Sensing Image Generation (DIRSIG) image simulation suite \cite{Goodenough2012DIRSIG5C}.
DIRSIG is a physics-based ray-tracing image simulation software suite that enables high-fidelity HSI simulations.
It can simulate complex light interactions within a scene using first-principles radiation propagation, enabling radiometrically accurate synthetic image generation, and has been used in numerous applications \cite{Carson2015SoilSS, Rengarajan2016ModelingOF, May2017VisibleAT, Archer2015EmpiricalMA, Bennett2014SSAMA, Gerace2012UsingDT}.
We base our simulations on the LWIR Mako sensor, which has 128 spectral channels ranging from $7.8-13.4\mu$m \cite{Hall2016MakoAT}.

\begin{figure*}[]
  \centering
  \begin{subfigure}[b]{0.32\textwidth}
    \centering
    \includegraphics[width=\linewidth]{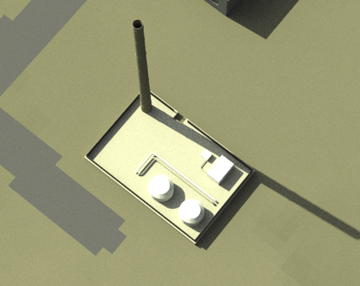}
    \caption{}
    \label{fig:facility}
  \end{subfigure}
  \hfill
  \begin{subfigure}[b]{0.32\textwidth}
    \centering
    \includegraphics[width=\linewidth]{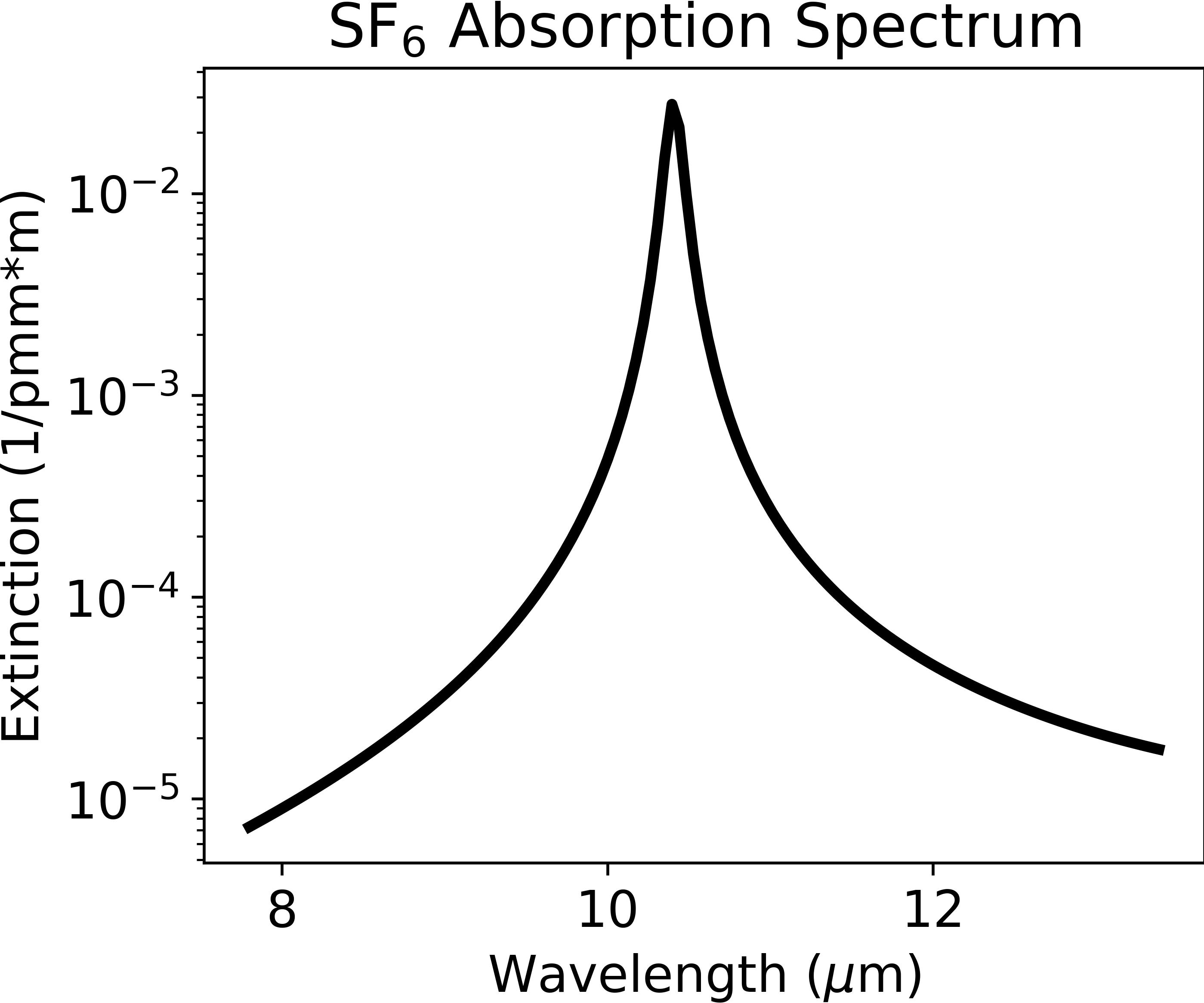}
    \caption{}
    \label{fig:gases}
  \end{subfigure}
  \hfill
  \begin{subfigure}[b]{0.32\textwidth}
    \centering
    \includegraphics[width=\linewidth]{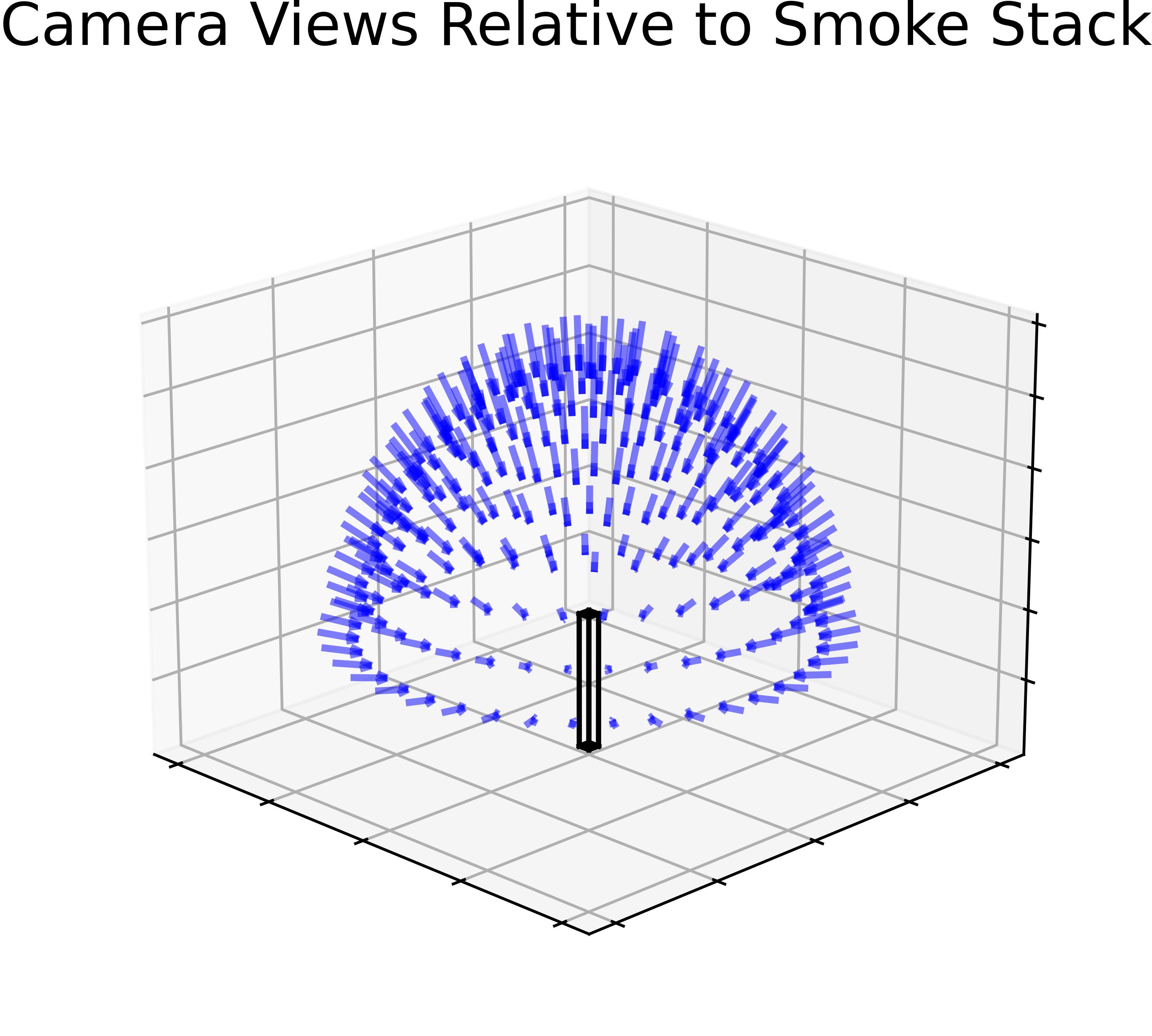}
    \caption{}
    \label{fig:coverage}
  \end{subfigure}
  
  \caption{(a) The visible light (RGB) coloring of our simulated scene, showing the stack, facility, road, and building. (b) The absorption spectrum of SF$_6$ used in the plume simulation. (c) The hemisphere of images captures the scene. Each image points to the base of the stack.}
  \label{fig:DIRSIG}
\end{figure*}

Figure \ref{fig:facility} shows a color rendering of our 3D scene, including a smokestack, roads, and buildings.
The plume is simulated using a version of the Blackadar plume model \cite{Blackadar1997TurbulenceAD}.
The plume is comprised of sulfur hexafluoride (SF$_6$), which has a strong and distinct absorption spectrum, with peak absorption near 10.5 $\mu$m as seen in Figure \ref{fig:gases}.
The plume was simulated at an emission temperature of 350K, a concentration of 110 ppm/s, a wind speed of 14.3 km/h, and a stack height of 38 m.
A total of 231 images were simulated, forming a hemisphere at a fixed distance of 1 km from the bottom of the stack, as seen in Figure \ref{fig:coverage}.
The scene and plume are intentionally simple to allow focus on developing the NeRF architecture and do not represent the full capabilities of DIRSIG to simulate highly realistic scenes.

We test training dataset sizes of 20, 30, 40, 50, 75, and 100 images to determine how NeRF performance is affected by the sparsity of training views.
For each training dataset size, five random image samples are tested.
For image samples, we do a biased farthest point sampling, where 65\% of the samples are taken from the top half of the hemisphere, with the remaining samples from the bottom half.
This sampling ensures uniform and sufficient coverage of the plume.
An evaluation dataset consists of a simple random sample of 31 images and is used to evaluate all models.
% All models were trained for 100,000 iterations with no early stopping.

\subsection{Image Reconstruction}

% This is the original "best" model numbers

% \begin{table}[ht!]
%     \centering
%     \caption{Average and standard deviation of PSNR and SSIM for Mip-NeRF and our method across different training set sizes. Our method consistently outperforms Mip-NeRF across all sizes.}
%     \begin{tabular}{lr|rr}
%         \textbf{Metric} & \textbf{\# Images} & \multicolumn{1}{c}{\textbf{Mip-NeRF}} & \multicolumn{1}{c}{\textbf{Ours}} \\
%         \hline
%         \multirow{4}{*}{PSNR $\uparrow$}
%             & 20 & 35.2 $\pm$ 0.14 & 37.4 $\pm$ 0.44 \\
%             & 30 & 36.0 $\pm$ 0.18 & 39.8 $\pm$ 0.67 \\
%             & 40 & 36.6 $\pm$ 0.49 & 41.3 $\pm$ 0.39 \\
%             & 50 & 37.2 $\pm$ 0.39 & 42.9 $\pm$ 0.30 \\
%             & 75 & 42.1 $\pm$ 0.65 & 45.5 $\pm$ 0.30 \\
%             &100 & 45.2 $\pm$ 0.20 & 47.2 $\pm$ 0.34\\
%         \hline
%         \multirow{4}{*}{SSIM $\uparrow$}
%             & 20 & 0.88 $\pm$ 0.004 & 0.90 $\pm$ 0.006 \\
%             & 30 & 0.89 $\pm$ 0.003 & 0.93 $\pm$ 0.007 \\
%             & 40 & 0.89 $\pm$ 0.005 & 0.94 $\pm$ 0.004 \\
%             & 50 & 0.90 $\pm$ 0.005 & 0.95 $\pm$ 0.003 \\
%             & 75 & 0.94 $\pm$ 0.004 & 0.97 $\pm$ 0.002 \\
%             &100 & 0.96 $\pm$ 0.002 & 0.98 $\pm$ 0.002 \\
%     \end{tabular}
%     \label{tab:recon metrics}
% \end{table}

For ease of analysis, we show here only how our model compares to the baseline Mip-NeRF model.
Comparisons between our model and MD and RegNeRF are in the Appendix.
We first evaluate how well the NeRF models can recreate the evaluation HSI.
To measure performance, we primarily consider the peak signal-to-noise ratio (PSNR) and the structural similarity index measure (SSIM) \cite{Wang2004ImageQA}, and report the average and standard deviation of each metric across five random samples for each method and training set size.

\begin{table}[b]
    \centering
    \caption{Average and standard deviation of PSNR and SSIM for Mip-NeRF and our method across different training set sizes. Our method consistently outperforms Mip-NeRF across all sizes. Larger numbers are better for PSNR and SSIM.}
    \begin{tabular}{lr|rr}
        \textbf{Metric} & \textbf{\# Images} & \multicolumn{1}{c}{\textbf{Mip-NeRF}} & \multicolumn{1}{c}{\textbf{Ours}} \\
        \hline
        \multirow{4}{*}{PSNR}
            & 20 & 33.7 $\pm$ 0.43 & 36.7 $\pm$ 0.77 \\
            & 30 & 34.5 $\pm$ 0.46 & 39.6 $\pm$ 0.72 \\
            & 40 & 35.4 $\pm$ 1.00 & 41.3 $\pm$ 0.41 \\
            & 50 & 36.4 $\pm$ 0.59 & 42.8 $\pm$ 0.33 \\
            & 75 & 42.0 $\pm$ 0.68 & 45.3 $\pm$ 0.32 \\
            &100 & 45.1 $\pm$ 0.20 & 47.1 $\pm$ 0.34 \\
        \hline
        \multirow{4}{*}{SSIM}
            & 20 & 0.86 $\pm$ 0.006 & 0.90 $\pm$ 0.009 \\
            & 30 & 0.87 $\pm$ 0.007 & 0.93 $\pm$ 0.007 \\
            & 40 & 0.89 $\pm$ 0.008 & 0.94 $\pm$ 0.004 \\
            & 50 & 0.90 $\pm$ 0.007 & 0.95 $\pm$ 0.003 \\
            & 75 & 0.94 $\pm$ 0.005 & 0.97 $\pm$ 0.002 \\
            &100 & 0.96 $\pm$ 0.002 & 0.98 $\pm$ 0.002 \\
    \end{tabular}
    \label{tab:recon metrics}
\end{table}

Table \ref{tab:recon metrics} presents the average metrics for both the standard Mip-NeRF and our proposed method using SAM and AWL2 losses with MD and GR.
We see that our method significantly outperforms Mip-NeRF for all training dataset sizes.
Our method achieves an average PSNR of 36.7 with 20 images, while Mip-NeRF requires 50 images to achieve a similar performance.
Thus, our method reduces the required number of training images by over one-half.
SSIM paints a similar picture, showing that the number of training images can be reduced by a factor of two.
However, with more images, both models tend to have similarly high performance.
With 100 training images, Mip-NeRF averages a PSNR of 45.1, while our method averages 47.1.

\begin{figure*}[b]
    \centering
    \includegraphics[width=\linewidth]{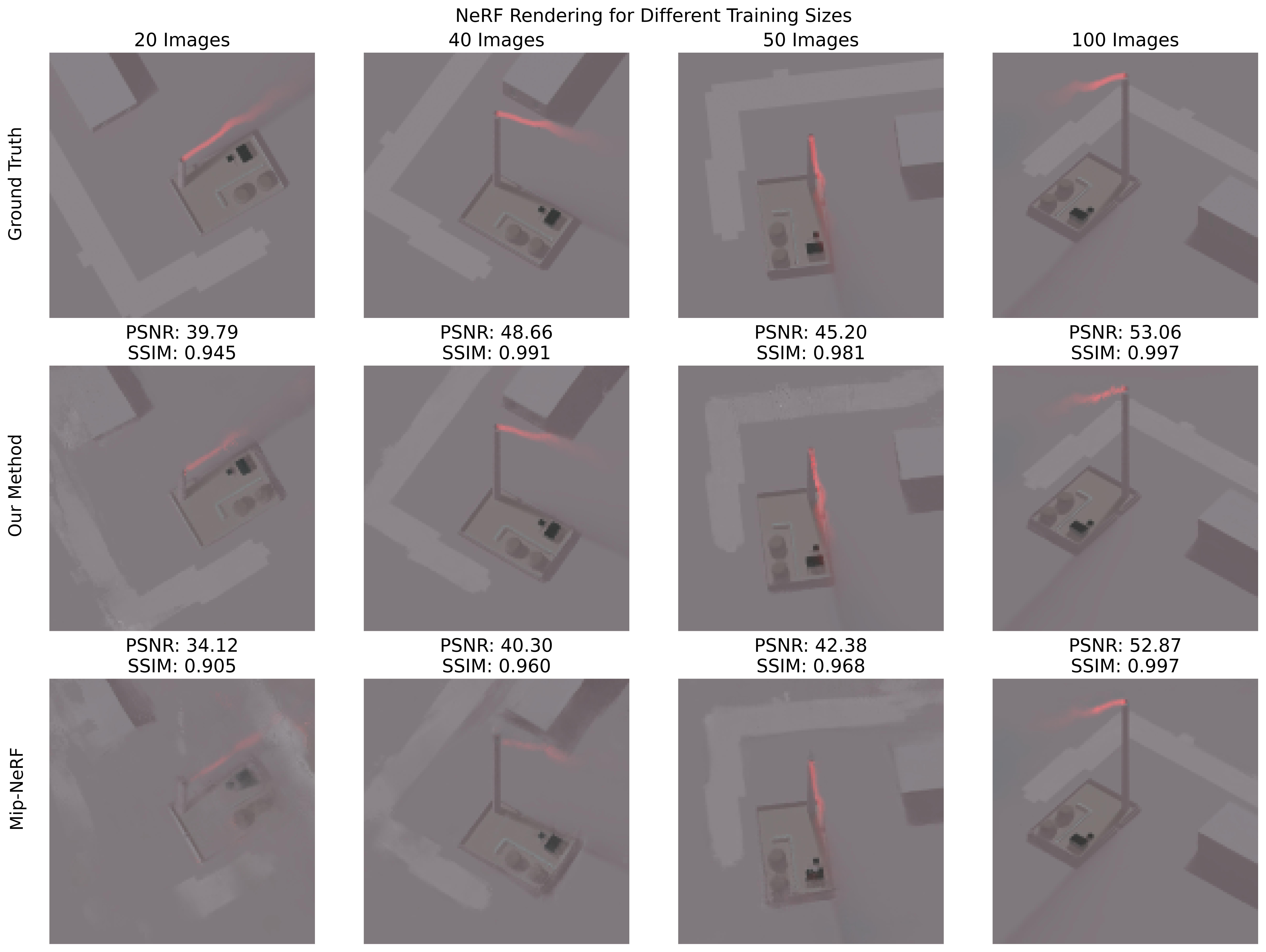}
    \caption{False-color renderings compared to ground truth images.
    The first row shows the ground truth, the second row shows the renderings from our method, and the third row shows the renderings from Mip-NeRF.
    The first column is for 20 training images, followed by 40, 50, and 100 training images.
    The seed that produced the highest geometric average of PSNR/55, SSIM, AUC, TPR, and 1-FPR for each image was used for rendering; these represent the best-case scenario for performance.
    The PSNR and SSIM are printed above each rendered image.
    See Video 1 (MP4, 4.7 MB) for ``drone path," false-color rendered videos of the eight models used to generate these still images.}
    \label{fig:recon examples}
\end{figure*}

Figure \ref{fig:recon examples} shows renderings from the NeRFs compared to the ground truth.
The seed that produced the highest geometric average of PSNR/55, SSIM, AUC, TPR, and 1-FPR for each image was used for rendering; these images represent the overall best-case scenario for each NeRF's performance.
With just 20 training images, Mip-NeRF can roughly capture basic scene features, but the results exhibit distortions and artifacts.
For example, the light gray road should be at the bottom of the scene, but it appears to be mingled with the plume.
Our method, however, is able to capture both the plume and actually capture the surrounding building and road geometry.
Doubling the number of training images to 40 improves the Mip-NeRF's ability to capture scene geometry, background, and plume features.
Qualitatively, this rendering looks similar to the rendering produced by our method with 20 images.
Meanwhile, our method trained on 40 images produces a rendering that visually matches the ground truth, though some of the fine details of the windows on the side of the building are blurry, and the sharp corners of the road are rounded.

With 50 training images, Mip-NeRF and our method visually produce similar results.
However, Mip-NeRF produces more geometric errors in the road and buildings than our method.
Additionally, the plume appears slightly fainter than the ground truth and our method's rendering, suggesting a potential underestimation of its radiance.
With 100 training images, both models produce visually accurate results, as confirmed by the average PSNRs of 45 and 47 for the Mip-NeRF and our method, respectively.
Our method's plume reconstruction still shows slight geometric errors, resulting in a plume that appears less smooth than the ground truth.

Video 1 shows video renderings from the models used to produce Figure \ref{fig:recon examples}.
In these videos, one can see that Mip-NeRF still produces occluding geometry with 20, 40, and 50 training images; only with 100 training images does Mip-NeRF produce a clean, consistent video render.
Our method produces exceptionally smooth, consistent geometry with only 40 training images.
With 20 training images, there are geometry mismatches, however they are greatly reduced in number, resulting in a significantly better reconstruction than Mip-NeRF.

Our method, both quantitatively and qualitatively, outperforms Mip-NeRF in capturing scene and plume geometry.
With few training images, our method is particularly performant compared to Mip-NeRF.
However, with many images, both models can produce similar results.
Training our method takes approximately twice the training time as for Mip-NeRF and requires about twice as much GPU memory.
A table of timing and GPU memory consumption values is available in the Appendix.
Using Mip-NeRF may be more feasible computationally, but it requires many more images to perform well, at least 100, according to our experiments.

\subsection{Gas Plume Detection}

We have shown that, with relatively few training images, our NeRF model can create a cohesive representation of an LWIR HSI scene and reconstruct the scene's spectral signatures from any reasonable viewing angle.
The next question is the potential utility of our model for gas plume analysis.
We consider here the first step in plume analysis: detecting the gas plume.
If we can satisfactorily detect the gas plume from unseen angles, this will suggest that the NeRF has learned sufficient radiometric information to improve gas plume analysis in future research.

% Next, we investigate the ability of novel rendered views from our NeRF model to be used for gas plume detection.
We apply ACE detection using the SF$_6$ absorption spectrum from Figure \ref{fig:gases} as $\mathbf{t}$ in Equation \ref{eq:ace}.
We then apply a threshold of 0.6 to the ACE detection map to create a binary plume mask, an example of which is shown in Figure \ref{fig:ACE example}.
The threshold value of 0.6 was found empirically to capture the majority of the plume with few false positives.
ACE detection is then applied to the NeRF rendered images, and the binary plume masks are compared.
The background distribution $\mu$ and $\Sigma$ are estimated from each individual image and are not the same between the ground truth and NeRF renderings.
The true positive rate (TPR) and false positive rate (FPR) are calculated for each test image and averaged to give overall model performance.
We additionally calculate the AUC for each image \cite{Hanley1982TheMA}.
The AUC can be interpreted as the probability that any given plume pixel has an ACE score higher than any given non-plume pixel.
An AUC of 0.5 would be equivalent to random guessing.

\begin{table}[]
    \centering
    \caption{Average and standard deviation of AUC, TPR, and FPR rates for Mip-NeRF and our method across different training set sizes. Our method often outperforms Mip-NeRF, particularly for 30 training images. Larger numbers are better for AUC and TPR, and smaller numbers are better for FPR.}
    \begin{tabular}{lr|rr}
        \textbf{Metric} & \textbf{\# Images} & \multicolumn{1}{c}{\textbf{Mip-NeRF}} & \multicolumn{1}{c}{\textbf{Ours}} \\
        \hline
        \multirow{4}{*}{AUC}
            & 20 & 0.588 $\pm$ 0.048 & 0.615 $\pm$ 0.076 \\
            & 30 & 0.638 $\pm$ 0.060 & 0.821 $\pm$ 0.081 \\
            & 40 & 0.750 $\pm$ 0.047 & 0.892 $\pm$ 0.041 \\
            & 50 & 0.832 $\pm$ 0.025 & 0.913 $\pm$ 0.009 \\
            & 75 & 0.940 $\pm$ 0.023 & 0.968 $\pm$ 0.013 \\
            &100 & 0.949 $\pm$ 0.026 & 0.987 $\pm$ 0.010 \\
        \hline
        \multirow{4}{*}{TPR}
            & 20 & 0.105 $\pm$ 0.033 & 0.214 $\pm$ 0.102 \\
            & 30 & 0.185 $\pm$ 0.086 & 0.557 $\pm$ 0.140 \\
            & 40 & 0.342 $\pm$ 0.118 & 0.642 $\pm$ 0.078 \\
            & 50 & 0.505 $\pm$ 0.074 & 0.705 $\pm$ 0.038 \\
            & 75 & 0.743 $\pm$ 0.036 & 0.818 $\pm$ 0.021 \\
            &100 & 0.778 $\pm$ 0.055 & 0.865 $\pm$ 0.017 \\
        \hline
        \multirow{4}{*}{FPR}
            & 20 & 0.004 $\pm$ 0.002 & 0.006 $\pm$ 0.001 \\
            & 30 & 0.006 $\pm$ 0.005 & 0.003 $\pm$ 0.001 \\
            & 40 & 0.004 $\pm$ 0.001 & 0.003 $\pm$ 0.001 \\
            & 50 & 0.003 $\pm$ 0.001 & 0.002 $\pm$ 0.001 \\
            & 75 & 0.003 $\pm$ 0.001 & 0.001 $\pm$ 0.000 \\
            &100 & 0.003 $\pm$ 0.001 & 0.001 $\pm$ 0.000 \\
    \end{tabular}
    
    \label{tab:detection metrics}
\end{table}

Table \ref{tab:detection metrics} gives the average AUC, TPR, and FPR rates averaged across the five random training seeds.
Our model outperforms the default Mip-NeRF across all training set sizes in AUC and TPR.
With 20 training images, Mip-NeRF achieves an AUC of 0.588, which is only slightly better than random guessing, whereas our method performs similarly and averages slightly higher.
Our model achieves an average TPR of 21.4\%, which is twice that of Mip-NeRF's 10.5\%.
This suggests that both models can begin detecting the gas with few images, but our method detects more plume pixels than Mip-NeRF.
With 30 images, our model's TPR jumps to 55.7\%, while the Mip-NeRF only increases to 18.5\% TPR.
Similarly, our model's AUC jumps to 0.821 while the Mip-NeRF increases to 0.638.
By 100 images, both models perform similarly in detection; however, our method still exhibits notably better performance.

Our method tends to have a lower FPR than Mip-NeRF.
However, the FPR rates for both models are exceptionally low, suggesting that false positives are unlikely to cause significant errors in plume analysis.
However, even with 100 training images, both methods have about an 80-85\% TPR, suggesting that, in conjunction with a small FPR, the plume rendered by NeRF tends to be slightly smaller than the true plume.
That is, NeRF accurately captures the bulk of the plume but struggles with its edges and fainter parts, producing only occasional extra plume pixels that result in false positives.
Thus, any estimates of plume volume from NeRF will likely be an underestimate of the true plume's size.

\begin{figure*}[]
    \centering
    \includegraphics[width=\linewidth]{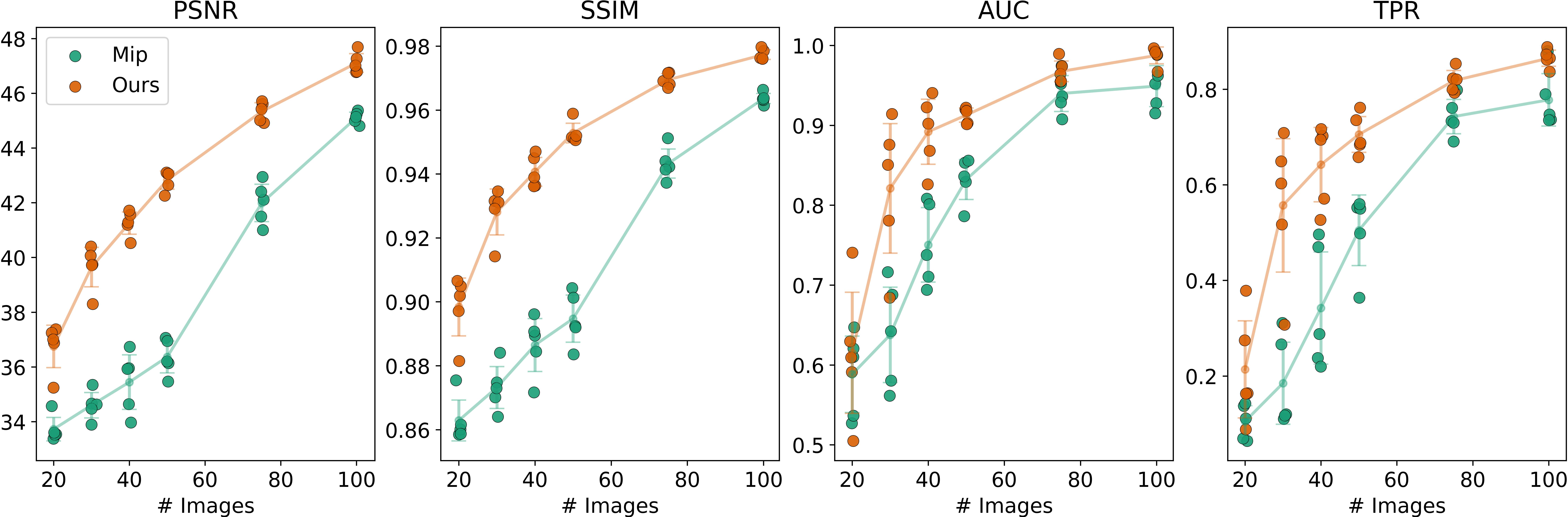}
    \caption{Plot of Mip-NeRF and our method's image reconstruction and detection performance (slight jitter applied to x coordinates). The lines and error bars show the average model performance, with standard deviation, from Tables \ref{tab:recon metrics} and \ref{tab:detection metrics}. Each of the five experiments is plotted to show the spread in model performance across different training samples. With the detection metrics AUC and TPR, a single experiment can produce high detection performance. This can be seen with our method on 20 training images, and with Mip-NeRF on 40 training images.}
    \label{fig:performance comparison}
\end{figure*}

Though average model performance shows slight improvements in detection when using our model, the average is hiding the range of performance across the five random samples.
Figure \ref{fig:performance comparison} shows the average and standard deviation for PSNR, SSIM, AUC, and TPR, along with the individual model metrics that were used to calculate the mean and standard deviation.
When looking at the detection metrics, we notice that one or two seeds pull our method's average performance up for 20 training images, while the other seeds perform about the same as Mip-NeRF.
This demonstrates the importance of testing different image samples if available, as well as potential different network initializations to obtain the best model performance, particularly with sparse training views.
We see similarly increased variability with Mip-NeRF, but with 30 and 40 training images.
Though our method has a seed that greatly improves detection performance, we do not see a similarly strong outlier in the PSNR and SSIM results with 20 training images.
This is likely because the plume occupies relatively few pixels in the scene.
Therefore, even partial reconstruction of the plume will lead to significant improvements in detection performance (which is entirely focused on the plume), but it will not necessarily improve PSNR, which is focused on the entire scene, not just the plume.
Additionally, it is curious that the random sample that produced high detection performance for our method did not produce good detection performance for Mip-NeRF.

When considering individual seed performance rather than the average across all seeds, we see that our method again significantly outperforms Mip-NeRF.
Table \ref{tab:min max metrics} shows the minimum and maximum metric values that each model achieved across the five random seeds.
We find that in PSNR and SSIM, our model's worst performance is always greater than Mip-NeRF's best performance, often by a significant margin.
For AUC and TPR, our model's worst performance is not always greater than Mip-NeRF's best performance, however the values are often very close.
Furthermore, when we consider our method's best performance, we see exceptional results that are always significantly greater than Mip-NeRF's best performance.
For example, our method achieves a max TPR of 37.9\% with 20 images, whereas Mip-NeRF requires 30 to 40 images to achieve similar performance.

\begin{table*}[]
\centering
\caption{Minimum and maximum model performance for all metrics. Values are organized as ``min, max''. These values can be found in Figure \ref{fig:performance comparison}. Our model's worst performance is frequently greater than Mip-NeRF's best performance. Larger numbers are better for PSNR, SSIM, AUC, and TPR. Smaller numbers are better for FPR.}
%%%%%%%%%%%%%%%%%%%%%%%%%%%%%%%%%%%%%%%%%%%%%%%%%%%%%%%%%%%%
% LEFT TABLE (top-aligned)
%%%%%%%%%%%%%%%%%%%%%%%%%%%%%%%%%%%%%%%%%%%%%%%%%%%%%%%%%%%%
\begin{minipage}[t]{0.48\linewidth}
    \vspace*{0pt} % forces top alignment
    \centering
    % \textbf{Image Reconstruction Metrics}\\[2mm]
    \setlength{\tabcolsep}{5pt}
    \begin{tabular}{lr|ll}
        \textbf{Metric} & \textbf{\# Img} & \textbf{Mip-NeRF} & \textbf{Ours} \\
        \hline
        \multirow{6}{*}{PSNR}
         & 20 & 33.38, 34.57 & 35.24, 37.37 \\
         & 30 & 33.89, 35.34 & 38.30, 40.40 \\
         & 40 & 33.97, 36.74 & 40.53, 41.72 \\
         & 50 & 35.47, 37.06 & 42.26, 43.10 \\
         & 75 & 41.00, 42.95 & 44.92, 45.71 \\
         &100 & 44.81, 45.38 & 46.78, 47.69 \\
        \hline
        \multirow{6}{*}{SSIM}
         & 20 & 0.858, 0.875 & 0.881, 0.907 \\
         & 30 & 0.864, 0.884 & 0.914, 0.935 \\
         & 40 & 0.872, 0.896 & 0.936, 0.947 \\
         & 50 & 0.884, 0.904 & 0.951, 0.959 \\
         & 75 & 0.937, 0.951 & 0.967, 0.972 \\
         &100 & 0.961, 0.966 & 0.976, 0.980 \\
    \end{tabular}
\end{minipage}
\hfill
%%%%%%%%%%%%%%%%%%%%%%%%%%%%%%%%%%%%%%%%%%%%%%%%%%%%%%%%%%%%
% RIGHT TABLE (top-aligned)
%%%%%%%%%%%%%%%%%%%%%%%%%%%%%%%%%%%%%%%%%%%%%%%%%%%%%%%%%%%%
\begin{minipage}[t]{0.48\linewidth}
    \vspace*{0pt} % forces top alignment
    \centering
    % \textbf{Detection Metrics}\\[2mm]
    \setlength{\tabcolsep}{5pt}
    \begin{tabular}{lr|ll}
        \textbf{Metric} & \textbf{\# Img} & \textbf{Mip-NeRF} & \textbf{Ours} \\
        \hline
        \multirow{6}{*}{AUC}
         & 20 & 0.527, 0.647 & 0.505, 0.741 \\
         & 30 & 0.561, 0.716 & 0.684, 0.914 \\
         & 40 & 0.694, 0.808 & 0.826, 0.941 \\
         & 50 & 0.786, 0.856 & 0.902, 0.922 \\
         & 75 & 0.908, 0.975 & 0.955, 0.989 \\
         &100 & 0.915, 0.988 & 0.967, 0.997 \\
        \hline
        \multirow{6}{*}{TPR}
         & 20 & 0.064, 0.143 & 0.088, 0.379 \\
         & 30 & 0.110, 0.311 & 0.307, 0.709 \\
         & 40 & 0.220, 0.496 & 0.527, 0.717 \\
         & 50 & 0.364, 0.560 & 0.658, 0.761 \\
         & 75 & 0.691, 0.799 & 0.793, 0.854 \\
         &100 & 0.736, 0.880 & 0.837, 0.888 \\
        \hline
        \multirow{6}{*}{FPR}
         & 20 & 0.002, 0.007 & 0.005, 0.008 \\
         & 30 & 0.002, 0.016 & 0.002, 0.005 \\
         & 40 & 0.002, 0.005 & 0.002, 0.004 \\
         & 50 & 0.002, 0.005 & 0.001, 0.003 \\
         & 75 & 0.002, 0.004 & 0.001, 0.001 \\
         &100 & 0.002, 0.003 & 0.001, 0.001 \\
    \end{tabular}
\end{minipage}
\label{tab:min max metrics}
\end{table*}

\begin{figure*}[t]
    \centering
    \includegraphics[width=\linewidth]{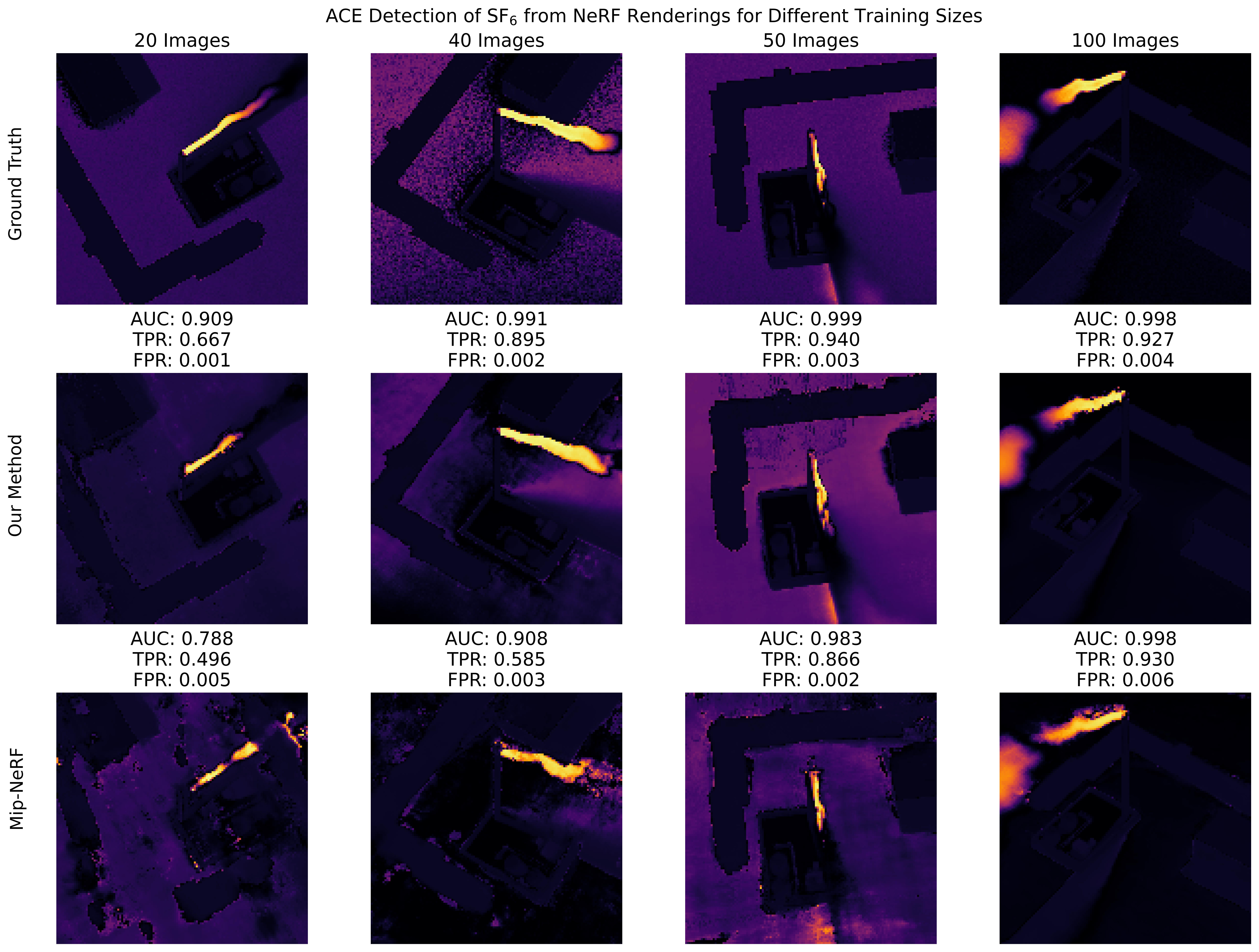}
    \caption{ACE detection scores for SF$_6$ applied to NeRF rendered images compared to ACE detection applied to the ground truth images. The rows show the ground truth, our method, and Mip-NeRF detection results. The columns show results for training set sizes of 20, 40, 50, and 100 images. These detections are applied to the same images as in Figure \ref{fig:recon examples}.
    The AUC, TPR, and FPR rates are printed above each reconstructed image.
    See Video 2 (MP4, 10.7 MB) for ``drone path," ACE rendered videos of the eight models used to generate these still images.}
    \label{fig:detection examples}
\end{figure*}

Figure \ref{fig:detection examples} shows ACE detection map examples using the best performing seeds for each model and image as described for Figure \ref{fig:recon examples}.
Looking at the 20-image NeRFs, we see that the plume is beginning to be captured by both models.
However, both models produce a plume that is overall smaller than the ground truth.
Additionally, Mip-NeRF has a higher false positive rate of 0.5\% compared to our 0.1\% FPR, and these false positives are easily seen at the tail of Mip-NeRF's plume.
With 40 training images, our method appears to enable easy plume detection and more accurate plume geometry compared to using 20 training images.
Furthermore, we notice that the ACE scores of the non-plume materials appear smoother and weaker than the ground truth.
Underneath the plume, there are slightly increased ACE scores from the plume's spectral shadow.
This suggests that our method can match the radiative physics demonstrated by the training images, whereas Mip-NeRF cannot.

With 50 training images, we notice differences between the two models.
Mip-NeRF appears to slightly more accurately detect the plume's size, while our method's overall detection scores are much smoother compared to Mip-NeRF.
With 100 training images, both our method and Mip-NeRF closely match the ground truth.
There are slight detection errors along the edge of the plume with our method.
Mip-NeRF also captures the bulk of the plume, however there are many more small detection errors around the edge of the plume.

We note that for both methods and the ground truth, there is a gap in the detection scores between the source and tail of the plume.
This corresponds to the plume's temperature cooling farther from the stack, at which point it matches the background temperature, resulting in a gap in the detection.
The plume then continues to cool, allowing it to be detected again.
This suggests that our NeRF can match and learn the plume's temperature from the training images.
However, more work is needed to investigate the accuracy of plume quantification using NeRF.

%%%%%%%%%%%%%%%%%%%%%%%%%%%%%%%%%%%%%%%%%%%%%%%%%%
\section{Conclusions}
\label{sec:Conclusions}
%%%%%%%%%%%%%%%%%%%%%%%%%%%%%%%%%%%%%%%%%%%%%%%%%%

% \begin{figure*}[ht!]
%     \centering
%     \includegraphics[width=\linewidth]{Figures/mini_Loss_Ablation.jpeg}
%     \caption{Performance comparison of Mip-NeRF when using different loss functions. These are the average and STD from 3 random initializations for each dataset size and loss function option. Including SAM produces the largest increase in performance, except for PSNR and SSIM, even with only 25 training images.}
%     \label{fig:mini loss ablation}
% \end{figure*}

% NeRFs make it easier to perform photogrammetry for HSI than traditional methods.
Capturing multiple hyperspectral images of a scene could offer greater opportunities for HSI remote sensing analysis.
Traditional photogrammetry methods have proved difficult to apply to HSI, yet the development of NeRFs has offered a promising alternative for 3D scene understanding of HSI.

We explored using NeRFs to learn a representation of a 3D scene from LWIR HSI data, with a particular focus on gas plume analysis.
The basic Mip-NeRF model can recreate HSI with sufficient accuracy for plume analysis when enough training images are available.
However, in practice, HSI views are likely to be limited.
Thus, we explored modifications to Mip-NeRF to increase performance with fewer images.

We first added a SAM loss function alongside the standard L2 loss.
We also developed an adaptive weighted L2 loss that further improved gas plume detection performance.
To further increase model performance with fewer training images, we combined multi-channel density NeRF with geometry regularization from RegNeRF.

Compared to regular Mip-NeRF, our method is able to achieve similar performance with 40-60\% fewer training images.
Our model achieves an average PSNR of 36.7 with 20 training images, while Mip-NeRF averages 36.4 PSNR with 50 training images.
Detection performance can depend on getting a good sample of images and the model initialization.
With 20 images, our model at its best achieves an AUC of just under 0.75 and a TPR of just under 38\%.
With 100 training images, both Mip-NeRF and our model perform similarly well, though our model still outperforms Mip-NeRF.

We have shown that it is indeed possible to use NeRF to create a cohesive representation of an LWIR HSI scene.
This representation combines all available information from the training images and allows HSI to be generated from any reasonable viewing angle.
We then showed that the radiometric information required to detect the gas plume remained present in the new renderings.
This is a promising result and suggests that NeRFs could be used to further enhance gas plume analysis in the future.
The main drawback is the number of available images.
We were able to create acceptable reconstructions with as few as 20 images.
However, even that many images may prove difficult to obtain in practice.
Additionally, the scene we simulated in DIRSIG was quite simple, and the additional complexity of real-world scenes will likely require more training images for NeRF to create a satisfactory reconstruction.

There are several options for future work.
The first is to further increase performance with fewer than 20 images.
One option is to investigate capturing a scene with both color and hyperspectral images and using the shared information to improve HSI reconstruction performance.
Another option is to further investigate the application of NeRFs for plume analysis.
Creating a 3D estimate of the plume's temperature and concentration, along with the plume's geometry, should be investigated.
Furthermore, additional plumes should be investigated, including weaker and more complex absorption features, weaker plumes that are more difficult to detect, and real-world captures of plumes.

\appendix

\section{Model Ablation Study, Model Hyperparameters, and Computation Time Comparison}
\label{app:ablation}

% Here is an ablation across sample sizes for going from MSE to L1+SAM, to multi-density, to geom-reg and annealing, to density L1 and AWL2.
All NeRF models and experiments were trained with the same hyperparameters and the same basic Mip-NeRF architecture.
The Mip-NeRF model had a base MLP with eight layers, 256 nodes wide.
The head MLP comprises two layers with 128 nodes each.
The integrated positional encoding used $L=16$ for the input encoding, and $L=4$ for the direction encoding.
We used $N_c=64$ coarse samples, and $N_f=196$ fine samples.
The networks were optimized using an Adam optimizer, with a learning rate that warmed up from $1e-5$ to $1e-3$ over the first 2000 iterations, then decayed back to $1e-5$ over 150,000 iterations, using a cosine ramp schedule.
All models were trained for 100,000 iterations.
Early stopping can be used and can improve NeRF performance.

We begin with an ablation study of the different loss function options.
We compare how Mip-NeRF performance changes when adding the additional loss functions.
The default for Mip-NeRF is just the L2 loss.
We consider adding the SAM loss with a weighting of two, as well as the AWL2 loss in addition to the SAM and L2 losses, with a scheduled $\lambda_{AWL2}$ as described in Section \ref{sec:Methods}.
We also investigate how performance changes for the MD and GR NeRF model when using just L2 loss, or L2 with SAM and AWL2.

%We begin with a small ablation study showing how performance changes when adding SAM and AWL2.
% Figure \ref{fig:mini loss ablation} shows image reconstruction and gas plume detection performance for the different loss options.
% The PSNR and SSIM plots show overall image reconstruction performance, while AUC and TPR show gas plume detection performance.
% We tested training dataset sizes of 25, 50, and 100 images, with three random initializations to get standard deviations.
% With just 25 images, all models perform similarly poorly, though using just L2 has a slight advantage.
% With 50 images, adding SAM markedly improved both image reconstruction and detection performance.
% Adding AWL2, in addition to SAM, offers slight performance improvements.
% This suggests that with a sufficient amount of images, adding SAM is beneficial, while adding AWL2 at least does not hurt performance.
% Furthermore, it appears that using just L2 loss is best with fewer images.

\begin{figure*}[t]
    \centering
    \includegraphics[width=\linewidth]{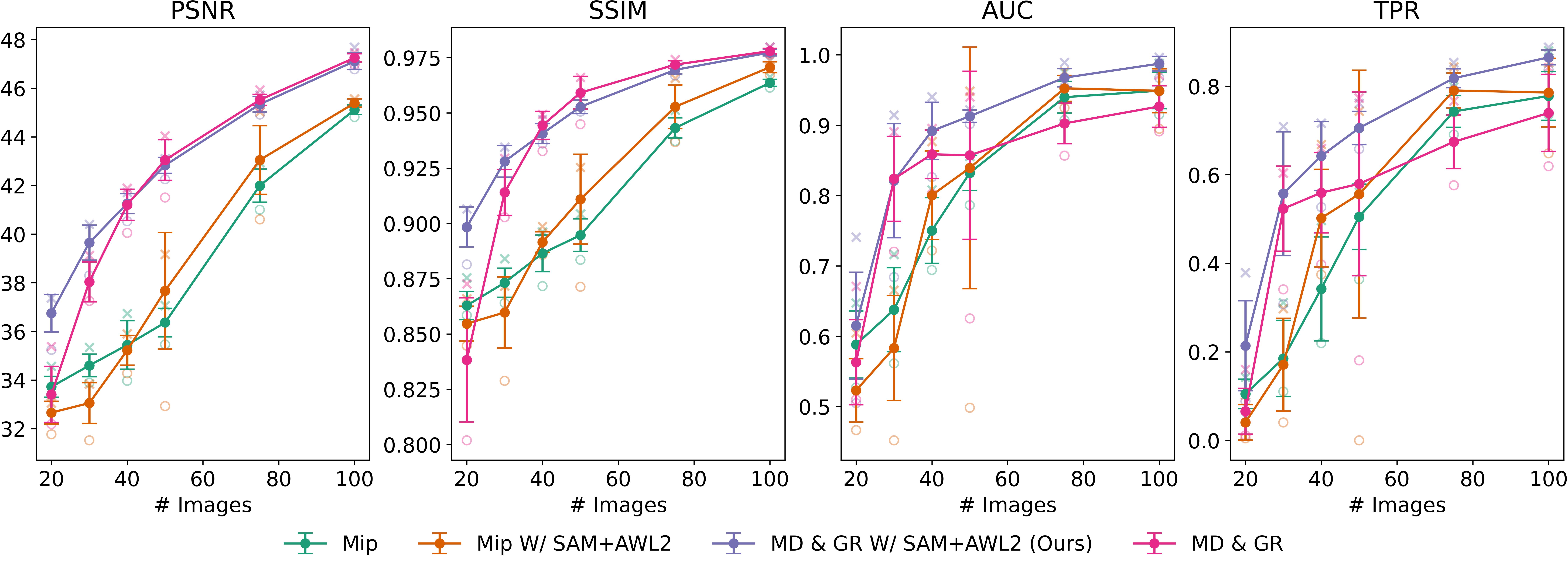}
    \caption{Image reconstruction and detection performance comparison when using just an L2 loss, and when adding SAM and AWL2. The green line shows default Mip-NeRF performance, while the orange line shows the change in performance when including SAM and AWL2. Additionally, the pink line shows performance with MD and GR, while the purple line shows performance with MD and GR when adding SAM and AWL2. The x's indicate the seed with the best model performance, and the o indicates the worst.}
    \label{fig:loss ablation}
\end{figure*}

Figure \ref{fig:loss ablation} shows our experimental results with training dataset sizes ranging from 20 to 100 images with five random initializations.
First, comparing default Mip-NeRF to Mip-NeRF with SAM and AWL2 loss, we see that for a few images, our proposed losses result in slightly worse image reconstruction and gas plume detection performance.
With 40 or more images, our proposed losses lead to improved performance across both image reconstruction and gas plume detection.

However, we do not see the same trend when testing different loss functions with our MD and GR NeRF model.
The pink line shows the MD and GR model performance trained using just an L2 loss.
The purple line shows the same model when including the SAM and AWL2 loss while training; this is what we call ``our" model in Section \ref{sec:Results}.
For image reconstruction performance, our proposed losses improve results, particularly with just 20 images.
With 40 images, using only the L2 loss function results in performance effectively the same as our method.
This is the opposite trend as we saw with the default Mip-NeRF model.
However, when considering detection performance, not using our additional losses results in consistently worse performance.
This suggests that the combination of our losses with MD and GR results in the best overall gas plume detection performance.
Particularly when only 20 images are available, removing any component of our proposed method will result in worse performance.
However, when many training images are available, say over 75, then using the default Mip-NeRF and only adding SAM will give much the same performance, with fewer computational requirements.

We next conduct an ablation study comparing the addition of MD and GR to standard Mip-NeRF.
For these experiments, we use the full three-part loss for all experiments.
We add MD by simply increasing the number of density outputs to 128 dimensions and adjusting the PDF sampling calculation.
We also consider adding GR by including a random patch depth smoothness loss and sample space annealing.
Annealing is used for the first 2000 iterations.
A total of 10,000 random cameras were generated, each with a resolution of $128\times128$ pixels.
Random patches of size $8\times8$ would be randomly selected from a patch during training.
A total of 16 patches are randomly sampled during each iteration.
When using GR and MD together, GR is applied to the average expected depth.

\begin{figure*}[]
    \centering
    \includegraphics[width=\linewidth]{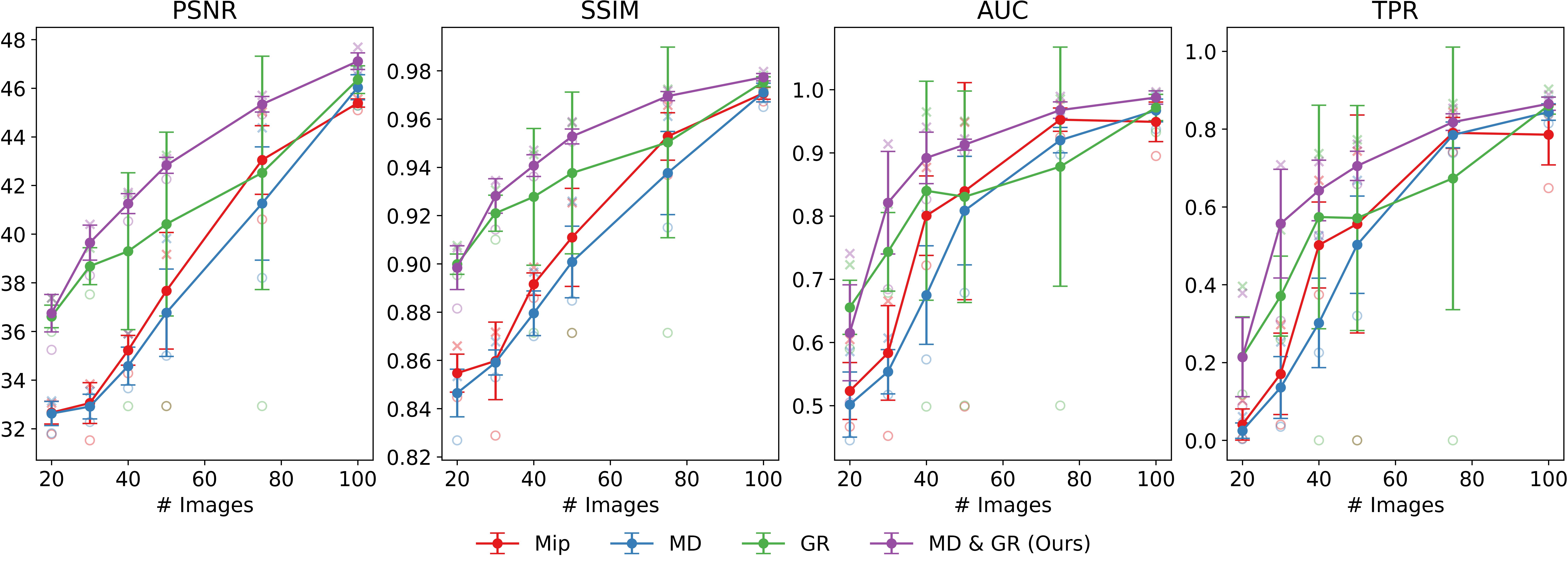}
    \caption{Performance comparison of various modeling changes compared to the default Mip-NeRF. All of these models use our proposed SAM and AWL2 losses. The red line shows the Mip-NeRF performance. The blue line shows performance when using multi-channel density. The green line shows the Mip-NeRF when adding geometry regularization and sample space annealing. The purple line shows the performance when combining MD and GR, and is the model we present as ``ours" in Section \ref{sec:Results}. The x's show the best-performing models, and the o's show the worst-performing models.}
    \label{fig:model Ablation}
\end{figure*}

Figure \ref{fig:model Ablation} shows the performance comparison of each model option.
The training dataset sizes are 20, 30, 40, 50, 75, and 100 images, with five random initializations for each model.
When comparing Mip-NeRF (red) to Mip-NeRF with MD (blue), we see that MD performs worse in both image reconstruction and plume detection, particularly for training sizes of 40, 50, and 75.
Comparing Mip-NeRF with added GR (green), there is a large performance increase in both image reconstruction and plume detection with up to 75 training images.
The number of training images required to match Mip-NeRF performance is roughly halved, up to 75+, after which all models perform roughly equally.
Combining MD and GR (purple, ``ours") yields improved image reconstruction performance compared to GR alone, particularly with 30 or more images.
In gas plume detection, our method has overall increased detection performance with 30 or more images.
There is increased variance across many of these experiments, driven by a few that perform exceptionally poorly.
This highlights the importance of not overtraining and exploring different model initializations and training image samples.

Our overall conclusion about which model additions are most beneficial depends on the number of training images.
With very few images (20-30), it appears that using all available options produces the best performance, but several different seeds and image samplings must be considered for the best performance.
For a medium amount of images (40-50), adding MD to GR does increase performance, however, at the cost of more training time and complexity.
Once many images are available (75+), standard Mip-NeRF with SAM can be used.
Not including GR, MD, or AWL2 reduces training time with negligible reduction in performance.

\begin{table}[]
    \centering
    \caption{Timing and GPU memory usage when training each method for 30,000 iterations, with 50 training images and 31 evaluation images.}
    \begin{tabular}{r|rr|rr}
         & \multicolumn{2}{c|}{Run Time} & \multicolumn{2}{c}{GPU Memory}\\
        Method & Time (s) & \% Inc. & Mem. (Gb) & \% Inc.\\
        \hline
        Mip-NeRF     & 3,190 &        & 13.0 &        \\
        Mip + Losses & 3,439 &  7.8\% & 13.0 &  0.0\% \\
        MD           & 4,339 & 36.0\% & 20.8 & 60.0\% \\
        MD + GR      & 5,525 & 73.2\% & 22.4 & 72.3\% \\
    \end{tabular}
    \label{tab:timing}
\end{table}

Lastly, we present training time and GPU memory usage for each model variation.
Time is the total runtime to train each model for 30,000 iterations, and GPU usage is the maximum memory used during training.
Table \ref{tab:timing} shows the runtime and GPU memory usage for default Mip-NeRF, Mip-NeRF with SAM and AWL2 losses, MD Mip-NeRF with additional losses, and our method, which is MD with GR and the additional losses.
In addition to raw runtime, we also include the percent increase in runtime and GPU memory compared to the default Mip-NeRF.

GPU memory usage will largely depend on the batch size (4096 rays) and the training and validation dataset sizes, which we set to 50 training images and 31 evaluation images for all models.
Furthermore, once AWL2 is added, the runtime will depend on how frequently the weights are updated and the number of training images.
When adding our additional SAM and AWL2 losses, the runtime increases by about 8\%, but GPU memory usage does not increase.
Adding MD to Mip, along with the additional losses, increases runtime by 36\% and uses an additional 60\% of GPU memory.
When adding GR to this model (``our" model), the run time further increases by 73\%, and the memory usage increases by 72\%.
Thus, we conclude that our additional losses only slightly increase run time, adding MD produces a large increase in GPU memory, and adding GR to MD slightly increases memory more than just adding MD, and it increases computation time a fair amount.

As previously discussed, the trade-off in computation time comes with the number of training images.
If few images are available, our method will produce better results, but at about a 70\% increase in computation time and memory usage.
However, if the training set comprises many images, standard Mip-NeRF can be used, reducing runtime and memory usage while achieving results similar to our model's.

% \subsection{Video}
% Acceptable file formats, including MOV (.mov), MPEG (.mpg), and MP4 (.mp4), are playable using standard media players, such as VLC or Windows Media Player. The recommended maximum size for each video file is 10-12 MB. Authors may insert a representative still image from the video file in the manuscript as a figure. The caption label will be linked by the publisher to the actual video file. The video may also be mentioned in an existing figure caption. Multimedia files are treated in the same manner as figures and they will be numbered sequentially with normal figures.  The video number, file type, and file size should be included in parentheses at the end of the figure caption. See Figure \ref{vid:satellite} for an example.

% \begin{video}
% \begin{center}
% {\includegraphics[height=5cm]{satellite.eps}}
% \\
% \end{center}
% \caption{\label{vid:satellite}This satellite is a still image from Video 1 (Video 1, MPEG, 2.5 MB).}
% \end{video}

% \disclosures 
\subsection*{Disclosures}
The authors declare that there are no financial interests, commercial affiliations, or other potential conflicts of interest that could have influenced the objectivity of this research or the writing of this paper

\subsection* {Code, Data, and Materials Availability} 
The code used to train and test the NeRFs, along with NeRF rendering functions and scripts, can be found on GitHub (\href{https://github.com/lanl/HSI-Nerfstudio}{https://github.com/lanl/HSI-Nerfstudio}).
Additional video renders to compare NeRF performance can be found on the GitHub page.
The DIRSIG simulated images used to train the NeRFs are available on Zenodo (\href{https://zenodo.org/records/18626884}{https://zenodo.org/records/18626884}).

\subsection* {Acknowledgments}
This work is funded by the FOCUS venture in the Remote Detection Portfolio of DOE/NNSA’s Office of Defense Nuclear Nonproliferation (DNN) Research \& Development.
It is an extension of previous work under the Living Backgrounds project, also funded through DNN R\&D. 
Generative AI was used during the code development process to help to write and debug functions.
LA-UR-26-20809.

%%%%% References %%%%%

\bibliography{bib}   % bibliography data in report.bib
\bibliographystyle{spiejour}   % makes bibtex use spiejour.bst

%%%%% Biographies of authors %%%%%

\vspace{2ex}\noindent\textbf{Scout Jarman} received a B.S. in Mathematics and Statistics Composite with a Computer Science minor from Utah State University in 2021.
He is pursuing a Ph.D. in Statistics \& Data Science from Utah State University with Dr. Kevin R. Moon as advisor.
He is conducting research in collaboration with the Intelligence and Space Research group at Los Alamos National Laboratory.
His research interests are in statistics and machine/deep learning applications to LWIR hyperspectral image analysis.

\vspace{1ex}\noindent\textbf{Zigfried Hampel-Arias} is a scientist in the Intelligence and Space Research Division at Los Alamos National Laboratory.
Dr. Hampel-Arias received his B.S. in Chemical Physics from Rice University and his Ph.D. in Particle Astrophysics from the University of Wisconsin-Madison.
They were a Postdoctoral Research Associate in the IIHE group at Universit\`e Libre de Bruxelles.
He has since directed his research interests to support national security efforts using machine learning and high-performance computing.

\vspace{1ex}\noindent\textbf{Adra Carr} holds a Ph.D in Experimental Condensed Matter Physics from the University of Colorado-Boulder and a B.S. in Physics from the University of Arizona.
Her background is in ultrafast laser spectroscopy and industrial materials science research, and she has worked on next-generation logic and memory technologies at IBM.
In her current position at Los Alamos National Laboratory, her research is focused on computational imaging and deep learning applied to hyperspectral imaging.

\vspace{1ex}\noindent\textbf{Kevin R. Moon} is an associate professor in Mathematics and Statistics and Director of the Data Science and Artificial Intelligence Center at Utah State University.
He earned B.S. and M.S. degrees in Electrical Engineering from BYU, and an M.S. in Mathematics and Ph.D. in Electrical Engineering from the University of Michigan.
Before joining USU in 2018, he was a postdoctoral scholar at Yale University.
His research spans machine learning, information theory, and their scientific applications.

\listoffigures
\listoftables

% \end{spacing}
\end{document}